\documentclass{article}

\pdfoutput=1

\usepackage[left=0.71in,top=0.95in,right=0.71in,bottom=0.75in]{geometry}
\usepackage{caption}
\usepackage{epsfig}
\usepackage{subfigure}
\usepackage{wrapfig}
\usepackage{sidecap}
\usepackage{amsmath}
\usepackage{setspace}
\usepackage{authblk}
\usepackage{diagbox}
\usepackage[title]{appendix}

\newcommand{\etal}{et al.\@}
\newcommand{\ie}{i.e.\@}

\begin{document}
\title{Early recurrence enables figure border ownership}
\author{Paria Mehrani and John K. Tsotsos}
\affil{Department of Electrical Engineering and Computer Science, \\
	York University, Toronto, Canada}
\affil{\{paria, tsotsos\}@cse.yorku.ca}
\date{}
\maketitle
\begin{abstract}
	The face-vase illusion introduced by Rubin demonstrates how one can switch back and forth between two different interpretations depending on how the figure outlines are assigned \cite{Rubin1915-illusion}. This border ownership assignment is an important step in the perception of forms. Zhou et al. \cite{BOwn_Zhou} found neurons in the visual cortex whose responses not only depend on the local features present in their classical receptive fields, but also on their contextual information. Various models proposed that feedback from higher ventral areas or lateral connections could provide the required contextual information. However, some studies \cite{BOwn_Craft, BOwn_Zhang2010, BOwn_Sugihara2011} ruled out the plausibility of models exclusively based on lateral connections. 
	In addition, further evidence \cite{Yau2012} suggests that ventral feedback even from V4 is not fast enough to provide context to border ownership neurons in either V1 or V2. As a result, the border ownership assignment mechanism in the brain is a mystery yet to be solved. Here, we test with computational simulations the hypothesis that the dorsal stream provides the global information to border ownership cells in the ventral stream. Our proposed model incorporates early recurrence from the dorsal pathway as well as lateral modulations within the ventral stream. 
	Our simulation experiments show that our model border ownership neurons, similar to their biological counterparts, exhibit different responses to figures on either side of the border.
\end{abstract}
\section{Introduction}
\begin{figure}[t]
	\centering
	\subfigure[][]{\label{subfig:fg_org_fig}\includegraphics[width=0.4\textwidth]{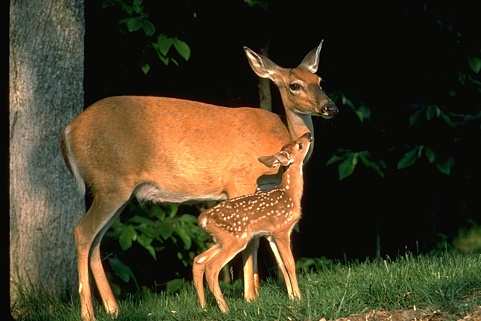}}
	\subfigure[][]{\label{subfig:fg_org_gt}\includegraphics[width=0.4\textwidth]{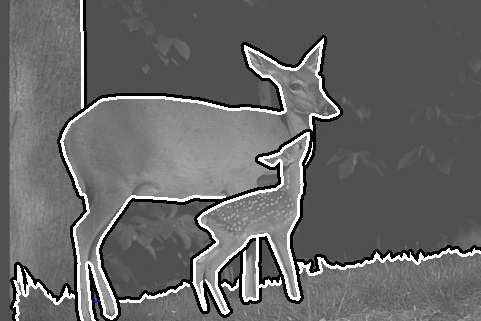}}
	\caption{\subref{subfig:fg_org_fig} An example of an image from the Berkeley dataset for ownership assignment at occlusion boundaries in natural images \cite{BSD_fowlkes2007local, BSD_ren2006figure}, \subref{subfig:fg_org_gt} the corresponding human-labeled ownership assignment along some of the occlusion boundaries. In \subref{subfig:fg_org_gt}, white represents the side that owns the border, \ie, side of the occluding object, and black is for the occluded region.}
	\label{fig:fg_organization}
\end{figure}

Objects present in a visual scene introduce occlusion borders by occluding either other objects or the ground. These occlusion borders are \textit{owned} by the closer object, \ie, the occluding one. Ownership information at occlusion boundaries plays a role for figure-ground assignment and perceptual organization. For example, in Figure \ref{fig:fg_organization}, an image and its corresponding hand-labeled ownership assignment along some occlusion boundaries are shown.
This ownership labeling provides a clear figure-ground organization of the observed scene, here, for example, the baby deer being in front of its mother.

\begin{figure}[t!]
	\centering
	\subfigure[][]{\label{subfig:BOS_bio_resp1}\includegraphics[width=0.45\textwidth]{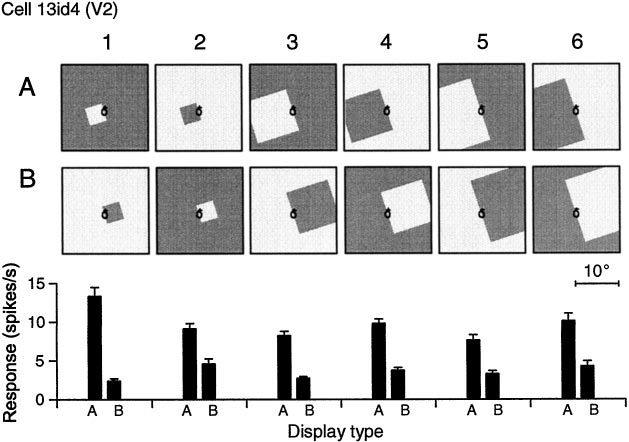}}
	\subfigure[][]{\label{subfig:BOS_bio_resp2}\includegraphics[width=0.5\textwidth]{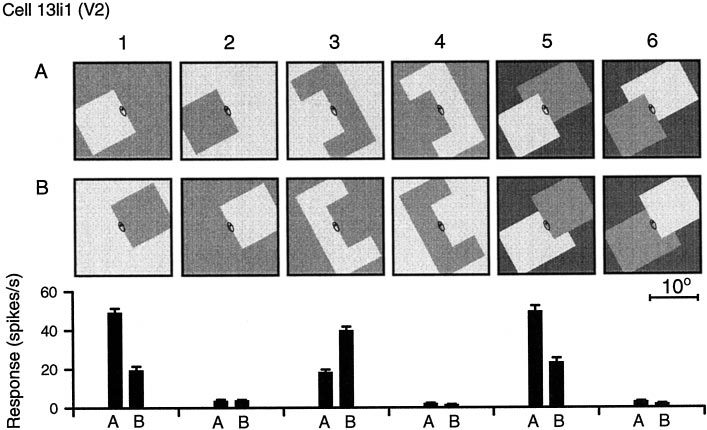}}
	\caption{Responses of type 1 \subref{subfig:BOS_bio_resp1} and type 3 \subref{subfig:BOS_bio_resp2} neurons. A black ellipse shows the classical receptive field of the recorded neuron in each display. The local features within the classical receptive field of each neuron are identical in each column of displays. The responses for both neurons, however, show a preference to stimuli with the figure on the left, while the type 3 neuron also exhibits selectivity to contrast polarity at the border (compare responses to displays in columns 1 and 2 of \subref{subfig:BOS_bio_resp2}). Although the large figures in \subref{subfig:BOS_bio_resp1} (columns 3-6) are not complete, according to the Gestalt principle of closure, these figures are to be perceived as square figures rather than holes with the rest of the image as the figure. Plots adapted from \cite{BOwn_Zhou}.}
	\label{fig:BOS_bio_responses}
\end{figure}

Although localizing the occlusion boundaries is a necessary step for figure-ground organization processing, determining ownership at these boundaries requires more than edge detection and local feature extraction. In fact, in a neurophysiological study, Zhou \etal \cite{BOwn_Zhou} showed that neurons exist in V1, V2 and V4, whose responses are context-dependent even with identical local features in their classical receptive fields across displays. As an example, consider the responses of a neuron from this study shown in Figure~\ref{subfig:BOS_bio_resp1}. In this figure, the little black ellipse in the middle of each display represents the classical receptive field of the neuron, and a close look reveals the same local features in pairs of stimuli A and B in each column. However, the context between the pairs differs in the sense that in displays of row A, the figure is on the left side of the border, while it resides in the opposite side for all displays in row B. The bars at the bottom of each column show the corresponding responses. For the neuron shown in this figure, the responses are stronger to stimuli in row A with the figure on the left and weaker to stimuli in row B with the figure on the right, for all pairs of displays presented. In other words, this neuron shows a preference to stimuli with the right side of the figure within its receptive field and a much weaker response to those with the left side of the figure in its receptive field. 
Zhou \etal \cite{BOwn_Zhou} called this type of selectivity the \textit{side-of-figure preference} and found that the context-dependent responses of these neurons encoded ownership information at occlusion boundaries and called them \textit{border ownership} cells.  

In their work, Zhou \etal \cite{BOwn_Zhou} identified four types of border ownership neurons. Type 1 neurons showed side-of-figure preferences, while their responses were independent of the contrast polarity on either side of the border. Responses of one such neuron are shown in Figure \ref{subfig:BOS_bio_resp1}. Type 2 neurons exhibited no side-of-figure preferences, but selectivity to the local contrast polarity, while type 3 cells were selective to both side-of-figure and contrast polarity in their receptive fields. Figure \ref{subfig:BOS_bio_resp2} demonstrates responses of a type 3 cell, where there are no responses to displays in even-numbered columns with local features not matching the selectivity of this neuron. The responses to stimuli in odd columns, however, show the preference for the figure to be on the left side of the border. Finally, neurons of type 0 demonstrated no obvious selectivity to either contrast polarity or side-of-figure features. Zhou \etal \cite{BOwn_Zhou} also found border ownership neurons maintain the difference of responses to figures of various sizes and changes in position of the border within their receptive fields. When presented with solid and outlined squares, border ownership responses were consistent across stimuli. As shown in the stimuli set for the type 3 neuron in Figure~\ref{subfig:BOS_bio_resp2}, the difference of responses was maintained for simple as well as more complicated stimuli, such as a C-shaped figure or overlapping squares. Interestingly, the average of responses to preferred and non-preferred stimuli (row A versus row B in Figure \ref{subfig:BOS_bio_resp1}, for example) over all cells showed divergence \textit{from the beginning} of stimulus onset, and the difference of responses achieved half-peak point at about 70 ms.

\begin{SCfigure}
	\centering
	\caption{Displays of fragmented Cornsweet figures for testing the effect of surround providing contextual information for border ownership assignments (adapted from~\cite{BOwn_Zhang2010}). The white dashed ellipse shows the receptive field of the recorded neuron. Squares on either side of the receptive field with both contrast polarities were tested. Each square is divided into eight fragments, including center and corner fragments of equal length. The center fragment on one side of the figure is placed on the receptive field. In each display, some of the remaining seven fragments are occluded, as some examples are shown in panel A. In a control study, the fragment centered on the receptive field is occluded, with some of the other fragments present (panel B). They also recorded the responses where the seven fragments of the square are scrambled (panel C).}
	\label{fig:Zhang_fragmented}
	\includegraphics[width=0.5\textwidth]{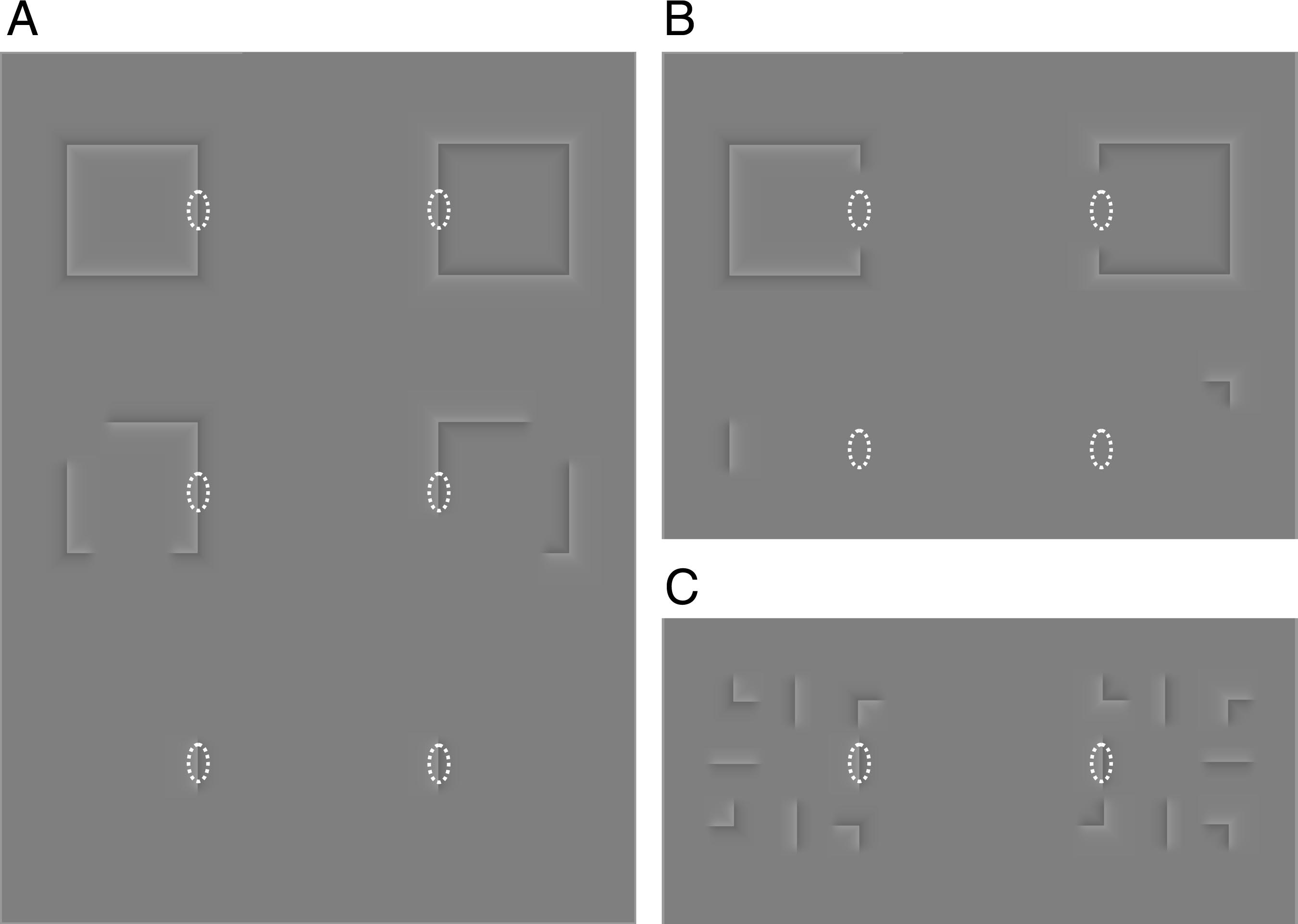}
\end{SCfigure}

Addressing the question of interactions between attentional modulation and border ownership assignment, Qiu \etal \cite{BOS_Qiu2007attention} observed that attention is not required for border ownership assignment and that border ownership was assigned for any figure in the display, both attended and ignored. However, they found that attentional enhancements were stronger for the figure on the preferred side of the border.
In an interesting work, Zhang and von der Heydt~\cite{BOwn_Zhang2010} investigated the structure of the surround that provides contextual information to border ownership neurons. Their stimuli set contained displays of fragmented Cornsweet squares\footnote{The contours of Cornsweet figures are step edges with exponential decays on either side such that the inside of the figure has the same intensity as the background \cite{Cornsweet}. Cornsweet stimuli allow for seamless occlusion of contour fragments.} such as those in Figure~\ref{fig:Zhang_fragmented}. While one edge was centered on the classical receptive field, some of the rest of the fragments were occluded, and the effect of the presence or absence of each fragment was studied.
They found all fragments have a modulatory effect on the response of the neuron in a uniform fashion, with positive and negative modulations from fragments of squares on preferred and non-preferred sides respectively. Furthermore, they obtained weaker border ownership signals for scrambled fragmented figures such as those presented in panel C of Figure~\ref{fig:Zhang_fragmented}.

Perhaps a prominent aspect of the study by Zhang \etal \cite{BOwn_Zhang2010} is the evaluation of the plausibility of a feedforward model with two ``hot-spots''\footnote{In their proposal, Sakai and Nishimura~\cite{BOwn_sakai2006surrounding} suggested employing modulatory signals from a facilitatory region and a suppressive region located asymmetrically on either side of an occlusion boundary for border ownership assignment. Zhang \etal \cite{BOwn_Zhang2010} called this proposal the two hot-spots (facilitatory and suppressive) hypothesis.} providing context to border ownership cells, similar to the one suggested by Sakai and Nishimura~\cite{BOwn_sakai2006surrounding}. They found that the population data does not confirm the assumption of such models. They examined the plausibility of models with only lateral connections\footnote{In Zhang's work \cite{BOwn_Zhang2010}, the term "lateral connections" stands for interareal interactions between neurons with receptive fields in neighboring regions of the visual field. In computational models such as convolutional neural networks (CNNs), lateral connections are implemented using  inter- and intra-feature-map interactions of neurons within a single layer with neighboring receptive fields. A depiction of such connections can be found in Figure \ref{fig:RL_connections}.}, such as that of Zhaoping's~\cite{BOwn_Zhaoping2005}. They discovered that lateral connections could not be the only source of contextual information for border ownership assignment and a model with both types of connections, feedback and lateral, could explain the data. Earlier, Craft \etal \cite{BOwn_Craft} had put forth a similar suggestion by providing a detailed discussion on the time course of border ownership cells from the study by Zhou \etal \cite{BOwn_Zhou} and speed of conduction for lateral connections. This suggestion was also confirmed in a later neurophysiological study by Sugihara~\etal~\cite{BOwn_Sugihara2011}, who tested border ownership neurons with displays of small and large figures. Sugihara \etal \cite{BOwn_Sugihara2011} found the latencies of border ownership cells increase for larger figures, yet are much faster to be affected by only lateral signal propagation. They argued that the latencies of border ownership neurons in their experiments did not rule out the possibility of a combination of feedback and lateral modulations.

Investigating the border ownership responses to occlusion boundaries in natural images, Williford and von der Heydt~\cite{BOwn_williford2016}, similar to previous studies, found that the selectivity of border ownership neurons originates from the image context rather than the contour configuration in the classical receptive fields. Strikingly, despite the complexity of natural images compared to the synthetic stimuli employed in previous work, the side-of-figure preference in these neurons started to emerge at about 70 ms, a similar time course observed with synthetic displays. Following this observation and considering the time course of IT cells reported by Brincat and Conner \cite{Brincat200617}, Williford and von der Heydt concluded that modulations from IT are not biologically plausible for border ownership assignment. In similar research, Hesse and Tsao~\cite{BOS_illusory2016} recorded border ownership neurons in V2 and V3. Their stimuli consisted of synthetic figures, intact natural faces as well as faces with illusory contours. In agreement with previous discoveries, these neurons showed side-of-figure preferences at around 65 ms, with increased latencies in case of stimuli with illusory contours.

Computational modeling of border ownership assignment was addressed in both computer vision and computational neuroscience communities. We will review both in Section \ref{sec:prev_work}. We also explore two feedback possibilities for border ownership assignment in Section \ref{sec:prev_work} and argue why one is not biologically plausible by analyzing the latencies reported in previous neurophysiological studies. In Section \ref{sec:our_model} our model architecture is described, as well as a detailed description of each layer in the hierarchy. Through simulation results in Section \ref{sec:results}, we demonstrate that our model border ownership neurons successfully mimic the behavior of the biological ones. Then, in sections \ref{sec:conclusion}, we conclude the work with some final remarks.

\section{Previous work}
\label{sec:prev_work}
The development of a computational model capable of determining border ownership at occlusion boundaries has been addressed by both computer vision scientists and computational neuroscientists, but often with different objectives. In computer vision, the goal is to have as accurate a border ownership label at occlusion boundaries as possible, by means of computational algorithms with no constraints on biological-plausibility. The performance in such models is usually measured against a  manually-labeled dataset. In contrast, the latter approaches aim at designing models that meet a set of known biological constraints while the simulated neurons have behavior similar to that of biological ones, hoping to suggest a mechanism for the emergence of the observed signal in the brain. Often, the performance of these models is judged on how well the simulated neurons replicate the behavior of their biological counterparts on a similar set of stimuli. Sometimes, the plausibility of these models is examined with further neurophysiological studies. In this section, we provide a review of both approaches.

\subsection{Computer vision models}
In computer vision, little attention has been devoted to this fundamental problem. Often, methods rely on motion and depth information for boundary assignments~\cite{BOwn_cv_motion_fu,BOwn_cv_motion_sundberg}. Inferring figure-ground organization in 2D static images was either addressed by limiting the input to line drawings~\cite{BOwn_cv_linedrawing_mumford, BOwn_cv_linedrawing_roberts}, or by extracting local features for natural images and applying machine learning techniques~\cite{BOwn_Hoiem2011, BOwn_cv_ren, BOwn_cv_fowlkes2007local}. Usually, approaches in the latter category formulate the problem in a conditional random field (CRF) framework to incorporate context into their model. A number of these models, such as~\cite{BOwn_Hoiem2011}, employ shape-related features like convexity-concavity of the contour segment or local geometric class like sky, ground, or vertical. Recently, a CNN-based model, called DOC, for figure-ground organization in a single 2D image was introduced~\cite{BOwn_cv_Wang2016}. Their network consists of two streams, one detecting edges and the other specifying border ownership, solving a joint labeling problem. They also introduced a semi-automatic method to provide figure-ground labeling for 20k images from the PASCAL dataset and showed superior performance over both PASCAL and figure-ground BSD~\cite{BSD500} dataset. 

Except for methods based on low-level cues, which are not among state-of-the-art, computer vision approaches generally attempt to solve the figure-ground organization problem by providing higher levels of features such as object class. For example, Hoiem~\etal~\cite{BOwn_Hoiem2011} first classified each superpixel in the image as one of vertical, sky, or ground, which provided strong priors over the object classes. In other words, these methods solve an object classification problem prior to addressing figure-ground organization. Even in the case of the deep model dubbed as DOC, although the network is trained on edge and occlusion boundary labels, there is no mechanism to ensure no implicit representation of object classes is learned. Given an object class, these models mainly leverage learned statistics of how likely an object is to be figure or ground. For example, the most likely event for the occlusion boundaries between any object and sky is to be owned by the object. 

To conclude, instead of solving the border ownership assignment before object recognition, these methods solve the latter first and infer border ownership based on learned statistics. This approach is indeed a detour to addressing the border ownership assignment problem and is in contrast to the findings about border ownership assignment. In particular, the short latencies of border ownership cells suggest that ownership assignment happens well before any shape processing in the ventral stream \cite{BOwn_Zhou, Yau2012} as we will discuss in detail in Section \ref{Section:Feedback}.

\subsection{Biologically-inspired models}
In the years following the border ownership study by Zhou \etal \cite{BOwn_Zhou}, there were a number of endeavors to suggest a mechanism for border ownership assignment in the brain. For example, Zhaoping~\cite{BOwn_Zhaoping2005} proposed a model of neurons in V1 and V2. Contextual information is provided by means of lateral connections in V2 and imposing Gestalt grouping principles such as continuity and convexity as well as T-junction priors. Sakai and Nishimura~\cite{BOwn_sakai2006surrounding} employed a vast pool of suppression and facilitation surround neurons for this purpose. In subsequent work, Sakai~\etal~\cite{BOS_sakai2012}, studied the responses of border ownership cells in this model to the complexity of shapes and found a decrease in responses with an increase in the complexity of shapes. Later, the plausibility of these models was ruled out by the neurophysiological findings of Zhang and von der Heydt \cite{BOwn_Zhang2010}. In the same year as Zhang and von der Heydt's work, Super \etal \cite{BOwn_Super2010} suggest another feedforward model based on surround suppression. An odd notion in their model is that feature maps for the figure and background are fed as input to their two-stream network: in a sense, they provide the ``answer'' in advance. 

The majority of computational models rely on feedback from higher layers. For example, Jehee \etal \cite{BOwn_Jehee2007} introduced a recurrent network with five areas V1, V2, V4, TEO and TE for contour extraction and border ownership computation. Similarly, border ownership assignment in models of Layton \etal \cite{Bown_Layton2014} and Tschechne and Neumann \cite{Neumann2014} is based on feedback from V4 and IT. Sugihara \etal \cite{BOwn_Sugihara2011} as well as Williford and von der Heydt \cite{BOwn_williford2016}, however, found that the time course of border ownership neurons does not support feedback from areas IT and beyond. 

Feedback from V4 in some of the biologically-inspired models in based on responses of grouping neurons. The earliest model, introduced by Craft \etal \cite{BOwn_Craft}, assigns border ownership using feedback from contour grouping cells in V4 and relies on convexity and proximity priors. Russell~\etal~\cite{BOwn_Russell} incorporated this model in a larger network and demonstrated improvements in saliency detection. Another extension of Craft's model \cite{BOwn_Craft} was introduced by Layton and colleagues~\cite{BOwn_Layton2012} with an additional layer of neurons, named R cells, with larger receptive fields than the grouping cells in V4. Then, border ownership is a result of feedback modulations from both grouping and R neurons, as well as imposed priors like convexity and closure. Despite the fact that no neurophysiological evidence against these models has been found, these approaches are too dependent on shape priors such as convexity and proximity. Moreover, they rely on feedback from grouping cells in V4 to provide the required contextual information. Nonetheless, as we discuss in detail in the next section, the time course of V4 neuron responses does not support the suggestion of border ownership modulations by feedback from these cells.

As a summary, the existing biologically-inspired models based on lateral connections or ventral feedback, though replicating the behavior of biological border ownership cells, cannot provide a timely signal carrying contextual information to border ownership neurons. As a result, the border ownership encoding process in the brain remains unclear.

\subsection{Exploring feedback possibilities}
\label{Section:Feedback}
Border ownership neurons reach half-peak strength for the difference of responses at about 69 ms and 68 ms in V1 and V2 respectively, and the divergence of responses to preferred and non-preferred stimuli was observed \textit{from the beginning} of stimulus onset \cite{BOwn_Zhou}. In this section, we examine two possibilities for providing the contextual information required for the exhibition of this divergence: first, ventral feedback from visual area V4, and second, dorsal modulations from MT. Neurons in both V4 and MT have receptive fields larger than those of V1 and V2 and could provide the required contextual information to border ownership cells. 

A number of studies showed that neurons in V4 exhibit selectivity to curvature~\cite{Pasupathy99_contour_features, pasupathy2002_nature, pasupathy2001_shape_representation}. Yau \etal \cite{Yau2012} examined the dynamics of curvature processing in V4 using a stimulus set of contour fragments, each with two orientation components forming various angles of curvature. Each contour fragment was presented at 8 orientations. Their contour tuning model consisted of linear and nonlinear terms; the linear term of the tuning accounts for the selectivity of each neuron to its orientation components, while the nonlinear responses represent shape (convexity/concavity) selectivity. In this study, they categorized cells as predominantly linear, predominantly nonlinear, and mixed linear/nonlinear neurons. A summary of the time course for each of the linear/nonlinear components of the tunings are shown in Table \ref{table:V4_time_course}.
\begin{table}
	\centering
	\begin{tabular}{|l||c|c|}
		\hline
		\diagbox{Cell Type}{Tuning Component}&\makebox[3em]{Linear}&\makebox[5em]{Nonlinear}\\\hline\hline
		Linear & $\sim$ 65 ms & $\sim$ 75 ms\\ \hline
		Mixed linear/nonlinear& $\sim$ 63 ms & $\sim$ 90 ms\\ \hline
		Average & $\sim$ 65 ms & $\sim$ 85 ms\\
		\hline
	\end{tabular}
	\caption{Time course of linear and nonlinear response components for predominantly linear and mixed linear/nonlinear cells in V4, as well as the average of all neurons (from Yau \etal \cite{Yau2012}, their Figure 5C, based on the time point of half-maximal signal strength). }
	\label{table:V4_time_course}
\end{table}
 Apparent from this table is that neither linear nor mixed linear/nonlinear neurons in V4 are fast enough to provide contextual information to border ownership cells in either V1 or V2, and this is the case for both linear and nonlinear components of responses. In other words, a feedback signal from V4 could not be the carrier of contextual information for the border ownership signal when it reaches its half-maximal point, let alone from the beginning of stimulus onset. The V4 cells, however, might be part of the mechanism that enhances the difference of responses to reach its peak or when there are ambiguities in the scene. In light of the time course of neurons in V4, models based on feedback from V4 described above are not biologically plausible, not only because even linear responses in V4 are not fast enough, but also due to the fact that these models rely on shape priors such as convexity, which emerge at a much later time in V4, after 75 ms.  

Another possibility for contextual modulation of border ownership neurons is MT in the dorsal pathway. A study on V4 and MT neurons by Maunsell \cite{Maunsell1987_timing} showed much shorter latencies in MT compared to V4, reaching half-maximal response strength at about 39 ms (their Figure 2.4, at half-maximal signal strength). MT cells not only have shorter latencies but also have large receptive fields, comparable to those of V4, which makes them perfect candidates to provide context to border ownership neurons. In fact, MT is one of the three regions that Bullier \cite{bullier2001integrated} called the ``fast brain'' with cells that are ``activated sufficiently early to influence neurons in areas V1 and V2''.  
It is worth emphasizing that the early divergence in the difference of responses in border ownership neurons could be attributed to the short latencies in MT cells. As a matter of fact, Hupe \etal~\cite{hupe1998corticalFeedback} showed that inactivating MT would affect neurons in areas V1, V2, and V3 to the extent that some neurons in these areas were silenced entirely in the absence of signals from MT. This effect was pronounced in the case of figure-ground stimuli, suggesting signals from MT contribute to figure-ground segregation. In another study, Hupe \etal~\cite{hupe2001feedbackTime} found that signals from MT significantly affected responses of V1 neurons \textit{early on} in the time course of responses and they concluded that these feedback connections are employed very early for the purpose of visual processing. Inspired by these observations, in the following section, we introduce a model of border ownership computation based on dorsal recurrent and ventral lateral modulations.

\section{Our Model}
\label{sec:our_model}
\begin{SCfigure}
	\centering
	\caption{Border ownership network architecture. Each scale, in both ventral and dorsal streams, defines an independent path from that of other scales. Border ownership neurons, in the ventral stream, receive a feedforward signal from complex cells, and a modulatory signal, showed with blue arrows, from MT neurons. In this network, scale selection is performed in the last layer, where border ownership neurons with the same selectivity across scales are combined. Finally, relaxation labeling, implementing ventral lateral connections, is performed on the neurons in the output layer to provide local context. Ventral lateral modulations happen between spatially neighboring neurons within a map as well as those across maps of similar local feature selectivities. An example of such ventral lateral connections is depicted in Figure \ref{fig:RL_connections}.}
	\includegraphics[width=0.5\textwidth]{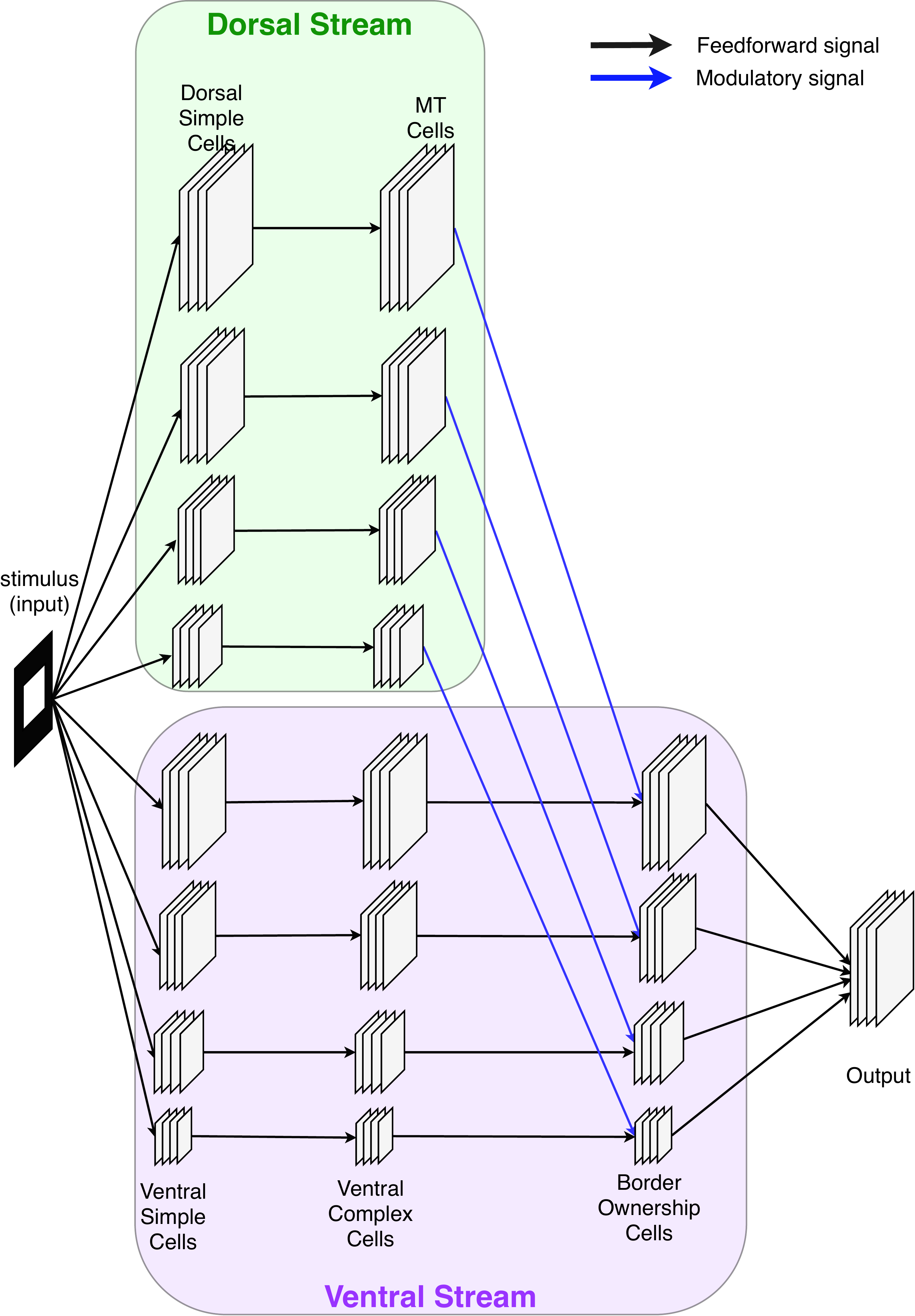}
	\label{fig:BOS_network}
\end{SCfigure}

\begin{SCfigure}
	\centering
	\caption{An example of the lateral connections for a single neuron, depicted as a big red circle, in the output layer of our model. The connections in this example are between neurons selective to dark bars on light regions at all orientations and ownership directions. The selectivity of each set of neurons is depicted in squares next to the set, with the red arrow indicating the ownership direction selectivity. Through iterations of relaxation labeling, these neurons interact with the big red neuron in excitatory or inhibitory manners and provide the local neighborhood consensus to this neuron. For a similar pictorial example of relaxation labeling for edge detection, we refer the reader to \cite{relaxation_zucker}, their Figure 10.2.}
	\includegraphics[width=0.4\textwidth,angle=90,origin=c]{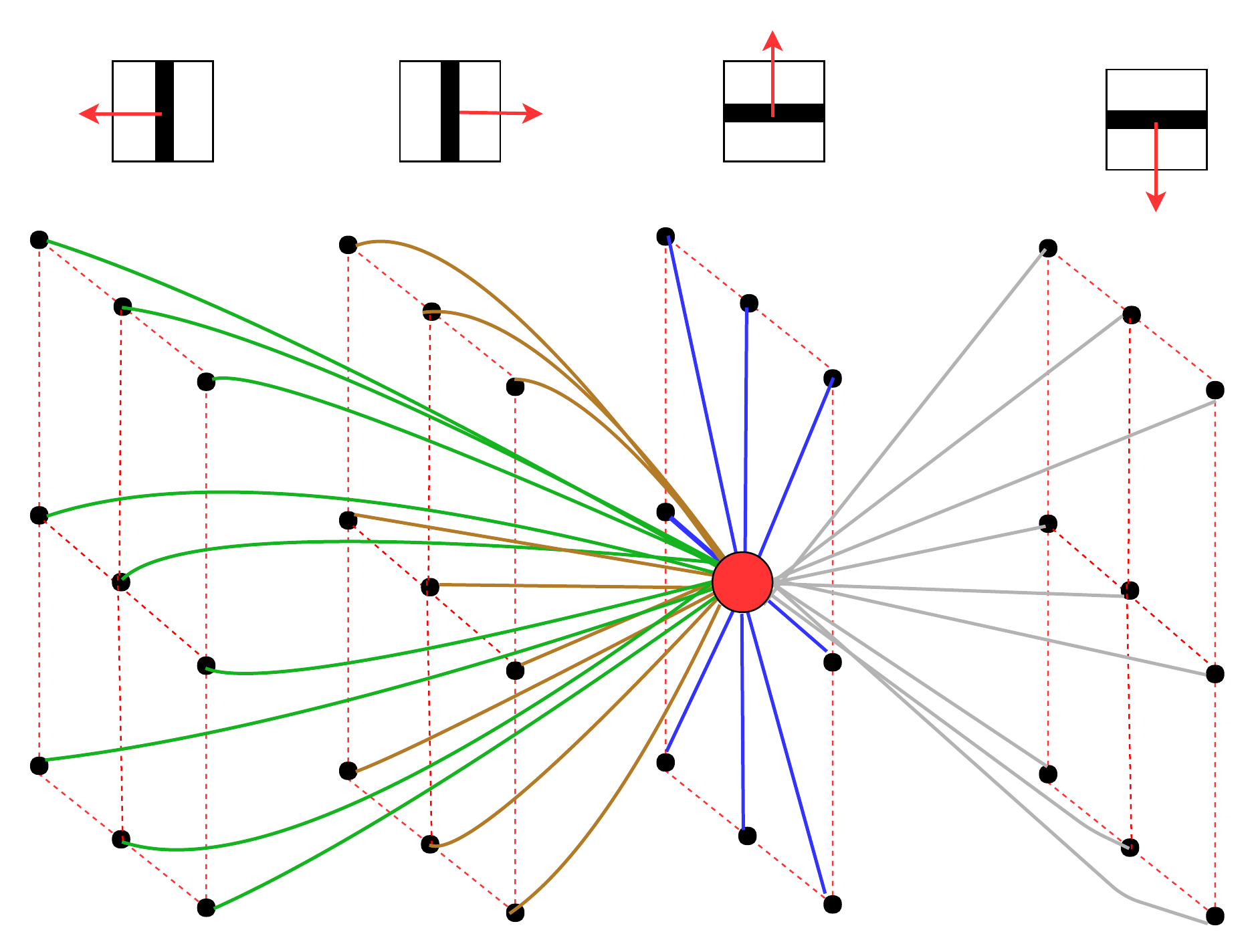}
	\label{fig:RL_connections}
\end{SCfigure}

In this work, we introduce a biologically plausible model for border ownership. Specifically, we introduce a model meeting known biological properties of the relevant brain regions, while suggesting a mechanistic computation of border ownership in the brain. Furthermore, a computer vision scientist can view our computational model as a convolutional neural network (CNN) \textit{with certain constraints}. The architecture of this network is depicted in Figure \ref{fig:BOS_network}, while Figure \ref{fig:RL_connections} represents an example of lateral connections in the output layer of this architecture. In the current work, horizontal and vertical orientations are implemented, with the goal of adding more variety of orientations to the model in future. Each layer of neurons is modeled at four spatial scales.

\begin{figure}[ht]
	\begin{minipage}[t]{.24\textwidth}
		\centering
		\includegraphics[width=\textwidth]{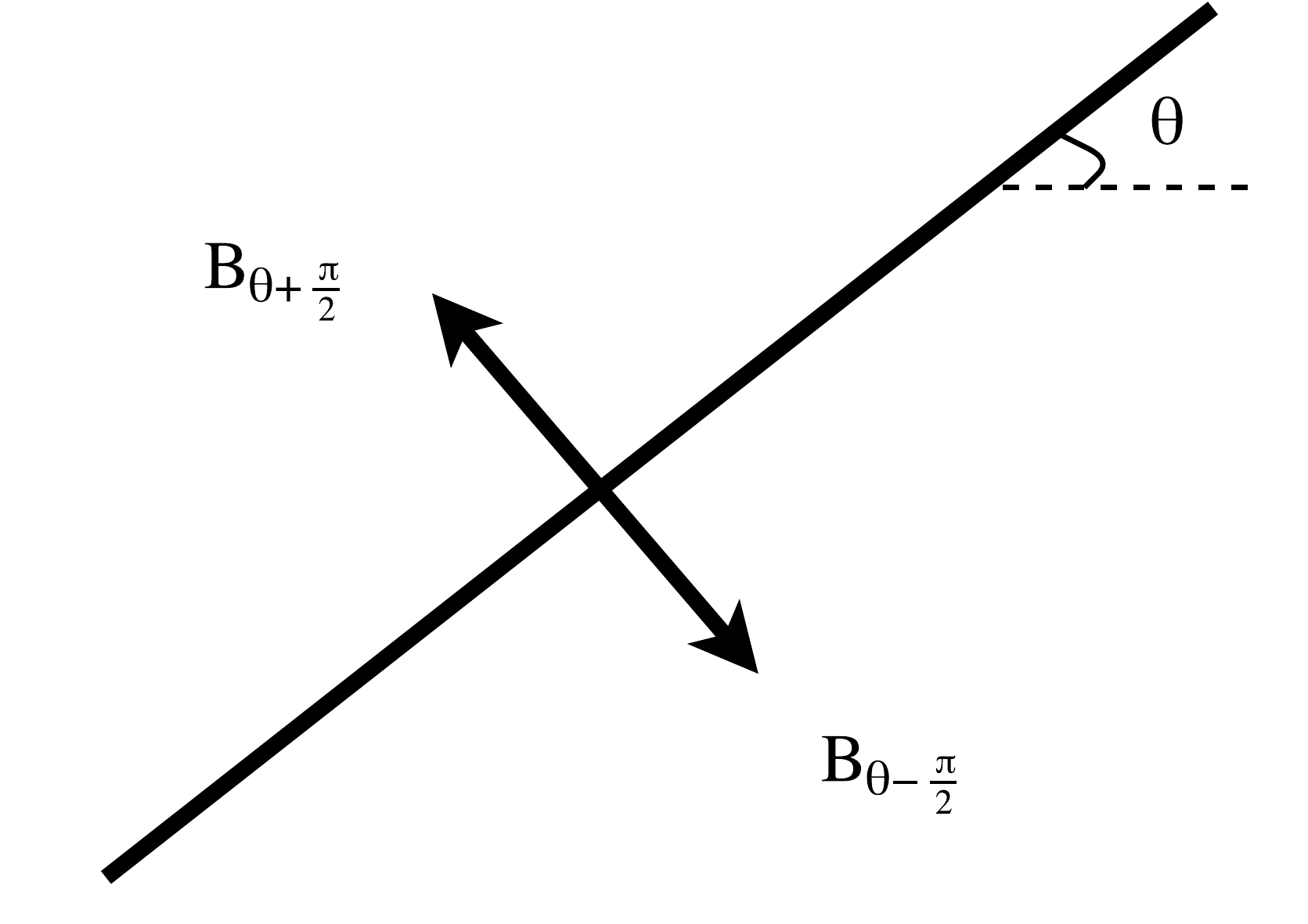}
		\caption{An example of an occlusion boundary oriented at $\theta$ degrees with two directions for two border ownership neurons each selective to figure on one side of the border.}
			\label{fig:border}
	\end{minipage}
	\hfill
	\begin{minipage}[t]{.7\textwidth}
		\centering
		\includegraphics[width=\textwidth]{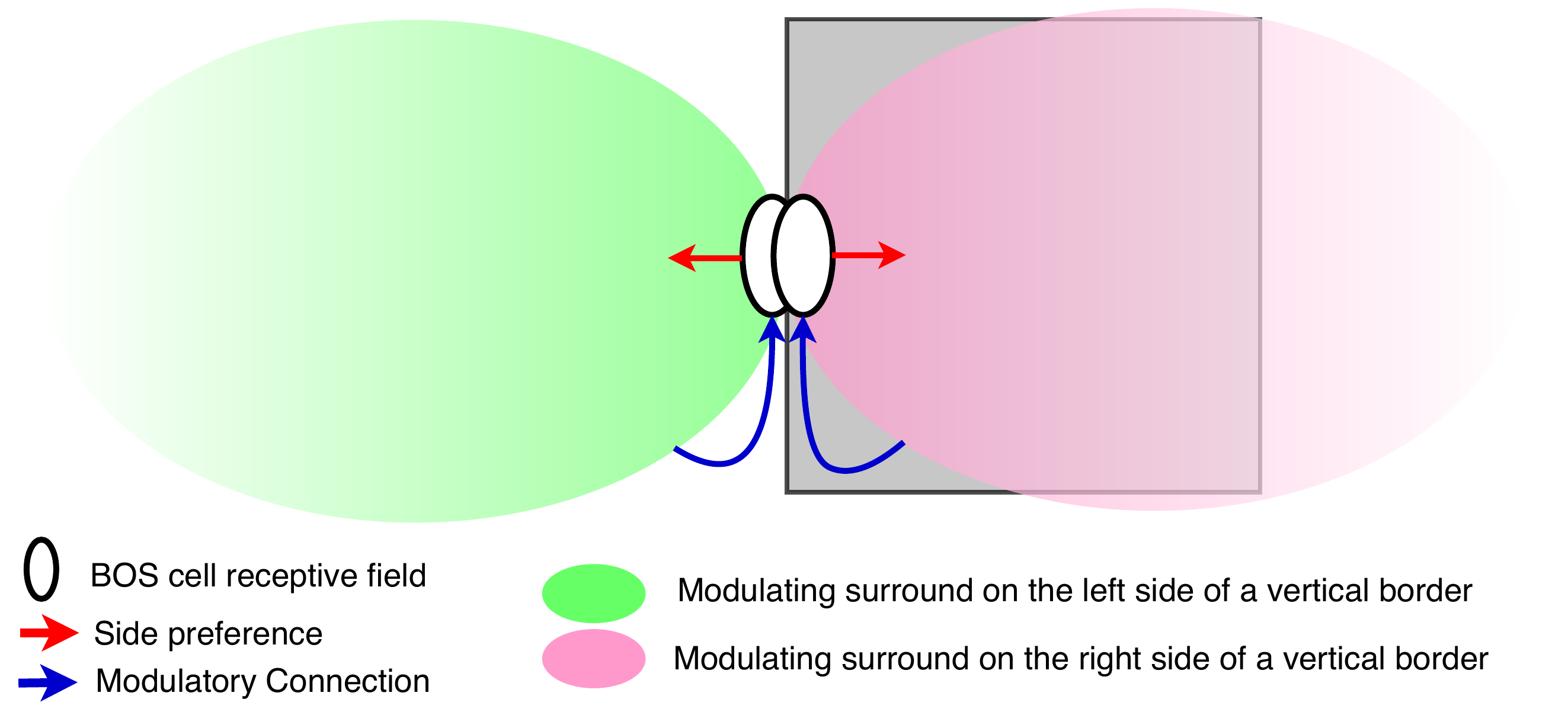}
		\caption{Two pools of border ownership cells for each visual field location encode the ownership at the border. A red arrow indicates the side preference of each BOS cell. Each border ownership cell receives a modulatory signal from MT cells on its preferred side. For example, the BOS cell with side preference to the right receives a modulatory signal from MT neurons with receptive fields in the pink surround area. These MT neurons notify the border ownership cell of prominent features on that side. The dorsal modulatory signal is computed as a Gaussian weighted sum of MT responses, which is indicated using a gradient filling in each surround area.}
			\label{fig:MT_modulations}
	\end{minipage}
\end{figure}

The goal of our work is to model border ownership neurons of types 1 and 3, classified by Zhou \etal \cite{BOwn_Zhou}. That is, neurons selective to side-of-figure and not contrast polarity (type 1) and those selective to both features (type 3). To this end, simple and complex cells in the ventral stream selective to contrast polarity and bars, from here on referred to as border- and edge-selective neurons respectively, encode local features as well as orientation within their classical receptive fields. These cells are responsible for signaling the existence of a border/edge to border ownership (BOS) cells. 
In our model, for each complex cell, border- and edge-selective alike, two border ownership neurons with the same orientation and local feature selectivity but opposite side-of-figure preferences are defined. For example, for a complex cell selective to edges at orientation $\theta$, we will have two border ownership neurons $B_{\theta + \frac{\pi}{2}}$ and $B_{\theta - \frac{\pi}{2}}$, one preferring the figure to reside on the right side of the edge and the other on the left side (see Figure~\ref{fig:border}). 
As a result, the final number of BOS cells for each visual field location is $2\times N\times S\times C$, where $N$ is the number of orientations, $S$ is the number of local feature selectivities in the model, and $C$ represents the number of scales. In our current model, we have $S=4$ for edge- and border-selective neurons each at two opposite contrast polarities, for example, borders between dark-light and light-dark regions, $N = 2$ for horizontal and vertical orientations, and $C = 4$. 

MT cells, carrying contextual information due to their large receptive fields, modulate border ownership neurons. Each border ownership cell, edge- or border-selective alike, receives the dorsal modulation from the MT cells on its preferred side. These MT cells notify the border ownership cell of the existence of prominent features on that side. For example, consider the BOS cell in Figure \ref{fig:MT_modulations} with figure preference on the right side, indicated by the red arrow. This neuron receives signals from the on- and off-center MT cells on the right side of the border, the surround area shown in pink. Most of the MT cells in the pink surround region are strongly activated due to the figure on that side and send a strong modulatory signal to the border ownership cell. In contrast, the modulation signal the BOS cell with the opposite ownership direction preference receives from the surround indicated in green is close to zero due to absence of conspicuous features.

The size of the surround determines the number of MT cells modulating each BOS cell. Moreover, the weight of the modulatory signal from each MT cell is computed as a Gaussian function of the distance between the receptive field centers of the MT and BOS cells. In other words, the closer an MT cell to a border ownership neuron, the stronger its modulatory effect on that neuron. In Figure \ref{fig:MT_modulations}, the weighting is reflected by a gradient filling in each surround region. Finally, after dorsal modulations, in the last layer of the network, scale selection over neurons with similar selectivities is performed.

Once an initial preference of border ownership is established, neighboring BOS neurons modulate each other by employing relaxation labeling~\cite{relaxation_zucker}. In particular, these lateral connections are another source for providing contextual information, this time, in the form of local context, and enforce collinearity for border ownership. In Figure \ref{fig:RL_connections}, the structure of lateral connections for relaxation labeling is presented. Each BOS neuron has a network of connections to other BOS cells at spatially neighboring visual field locations. In such a network of connections, when a BOS neuron has a strong response in one direction while its immediate neighbors believe otherwise, the confidence of its response is decreased and corrected over relaxation labeling iterations.  

This type of computation providing local context employs lateral connections between BOS neurons, but cannot be the only source of contextual information for border ownership assignments~\cite{BOwn_Zhang2010}. Nevertheless, an increase in latencies for larger figures observed by Sugihara \etal \cite{BOwn_Sugihara2011} could be described by a time-limited lateral modulation. As a summary, the final BOS activations are obtained from a combination of feedforward signals as well as dorsal and ventral lateral modulations. In what follows, we describe each cell type employed in our model in detail.

\vspace{15pt}
\textbf{Ventral simple and complex cells.} Following Rodr{\'\i}guez-S{\'a}nchez and Tsotsos \cite{Antonio_2012}, we implemented ventral simple cells with edge selectivity using difference of Gaussians (DoG). For parameter values of DoGs, we refer the interested readers to \cite{Antonio_2012}. Border-selective simple cells in the ventral stream were implemented using Gabor filters,
\begin{equation}
g(x, y; \lambda, \theta, \psi, \sigma, r) = \exp{\left(-\frac{x'^2 + r^2 y'^2}{2\sigma^2}\right)}\cos\left(2\pi\frac{x'}{\lambda}+\psi\right)
\end{equation}
with $r=0.5, \lambda=0.2$, $\psi = \pm \frac{\pi}{2}$ for two settings of contrast polarity, and $\sigma = \frac{\text{RF}}{4}$, where RF is short for receptive field size.  We measured 1$^\circ$ visual angle at 50 cm to be 32 pixels for stimuli size of $400\times400$ pixels, and set the ventral simple cell receptive fields at four scales to $[0.4^\circ, 0.6^\circ, 0.8^\circ, 1^\circ]$, following the observations of \cite{Angelucci2002cRF, Gattass1987V1RF}. Complex cells, as in \cite{Antonio_2012}, were computed by a Gaussian-weighted sum of simple cells. Both types of cells are half-wave rectified. In our implementation, all receptive fields are square-shaped, even though round shapes might be used in some of the figures in this document in a figurative manner.

\vspace{15pt}
\textbf{Dorsal simple cells and MT cells.}
The dorsal pathway is well-known to be selective to spatiotemporal features. However, a number of neurophysiological studies \cite{MT_static_albright1984direction, MT_static_Raiguel1999response, MT_static_kolster2010retinotopic, felleman1984MTreceptive} provided evidence that the dorsal area MT also responds to static stimuli. In our implementation, we skip the temporal aspect of dorsal responses as our input to the network are static images. But such an extension is straightforward.

Angelucci \etal \cite{Angelucci2002cRF} found V1 cells receiving signal from Magnocellular (M) and Parvocellular (P) cells in the LGN have different receptive field sizes. Neurons on the M-path have larger receptive fields and as a result, our dorsal simple cells have receptive field sizes $[0.9^\circ, 1.33^\circ, 1.76^\circ, 2.2^\circ]$. In our implementation, we set the parameters of edge-selective dorsal simple neurons as: $\sigma_y = \text{RF}, WR = 2.5, AR = WR * [10, 9, 8, 7]$, where $WR$ is the width ratio and $AR$ the aspect ratio of the DoG kernel (See \cite{Antonio_2012} for details). For the border-selective dorsal simple cells, the difference of two Gaussians with $\sigma_y = \text{RF}, \sigma_x = \frac{\sigma_y}{AR}$ were employed. At each visual field location, the dorsal simple cell with maximum activity feed to MT cells.

\begin{SCfigure}
	\centering
	\subfigure[on-center]{\includegraphics[width=0.3\textwidth]{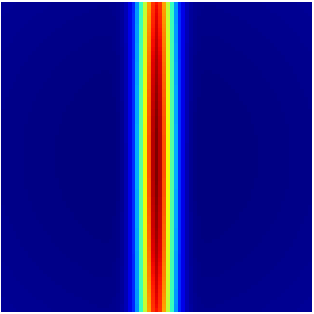}\label{subfig:MT_on_center}}~
	\subfigure[off-center]{\includegraphics[width=0.3\textwidth]{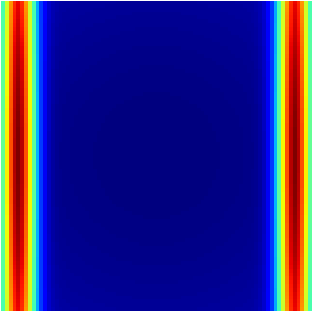}\label{subfig:MT_off_center}}
	\caption{Receptive fields of on-center and off-center MT cells. The on-center MT neurons have an on area in the middle (red color) and two off regions on the sides (blue color). The kernel of off-center MT cells comprises of two excitatory (red color) Gaussians with same parameters centered at the two sides of the kernel and one Gaussian with inhibitory effects (blue color) at the center.\vspace{1.3cm}}
	\label{fig:MT_profile}
\end{SCfigure}

The receptive fields of MT neurons were determined based on the findings of Fiorani \etal \cite{Fiorani1989MT}, set to $[2.5^\circ, 3.26^\circ,$ $4.02^\circ, 4.78^\circ]$. Felleman and Kass~\cite{felleman1984MTreceptive} found that MT cells, with on and off excitatory regions, showed an increase of responses to bars with the length up to the receptive field size. However, the effective bar width is $\frac{1}{10}$-th of the receptive field size. In fact, for MT neurons with 4$^\circ$ receptive fields, the most effective bar width was about 0.25$^\circ$. In other words, MT cells are selective to long narrow bars, with length matching the receptive field size. The receptive field of an example model MT cell with such selectivities and an excitatory region in the middle is depicted in Figure \ref{subfig:MT_on_center}. Such a receptive field profile is not unusual in the visual system as a similar one was suggested in cat simple cells by Hubel \cite{hubel1963visual} (their ``SIMPLE CORTICAL CELLS'' figure, subfigures a and b). We called these MT cells with the on area in the middle as ``on-center'' MT neurons and implemented these cells using the Difference of Gaussians formulation. We also implemented MT cells with an inhibitory region in the middle and two excitatory areas on the sides, similar to Hubel's subfigure f in his ``SIMPLE CORTICAL CELLS'' figure. An example of a model off-center MT neuron receptive field is shown in Figure \ref{subfig:MT_off_center}. Note that these neurons exhibit selectivity to long and narrow bars on either side of the receptive field with strong activations when two bars are present on both excitatory areas. We refer to these neurons as ``off-center'' MT cells, which were computed by the difference of three Gaussians, two with same parameters centered at the two sides of the receptive field, and one centered in the middle. While the activation of on-center MT cells signifies the existence of edges, off-center MT neurons indicate a pair of edges on either side of their receptive fields. As a result, the sum of off-center MT neurons at all orientations will be large for closed shapes, a key feature for border ownership assignment.
 For MT cells, we set $\sigma_y = \text{RF}$, $AR = [33, 52, 80, 126.6]$, and $WR= [3.3, 5.2, 8, 12.6]$ for the four scales of receptive fields.

\begin{figure}[t!]
	\centering
	\subfigure[][]{\label{subfig:MT_contrast_resp}\includegraphics[width=0.47\textwidth]{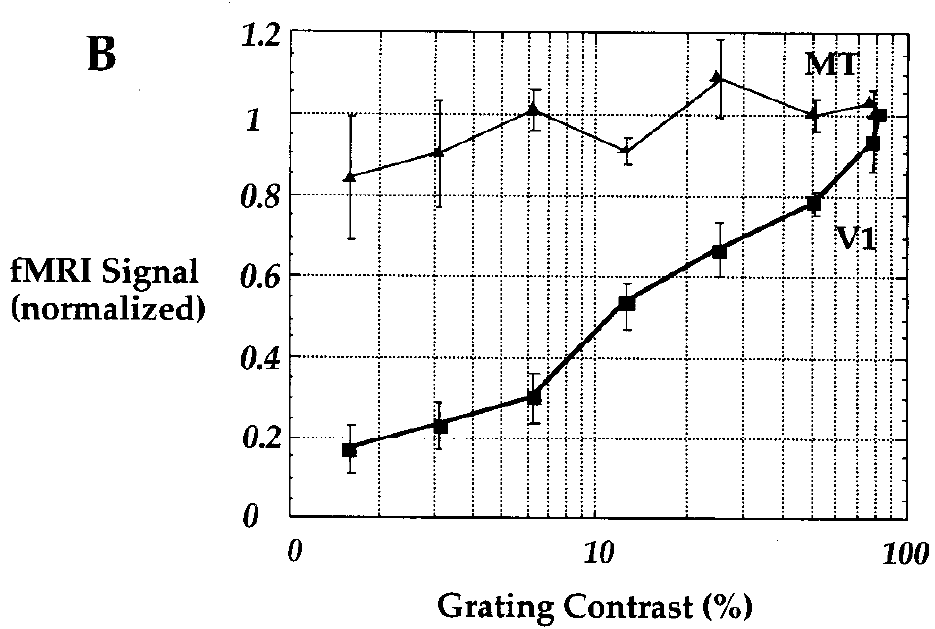}} ~
	\subfigure[][]{\label{subfig:phi_rectifier}\includegraphics[width=0.45\textwidth]{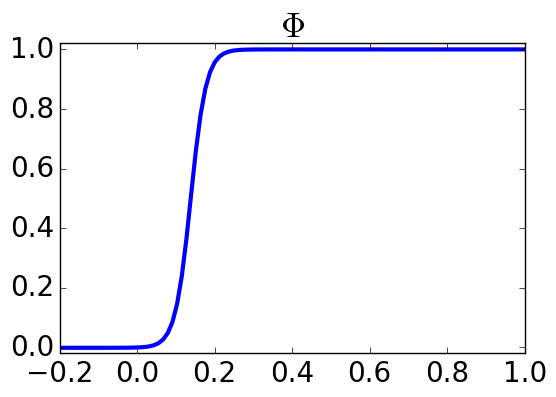}}
	\caption{MT cells show high sensitivity to contrast. \subref{subfig:MT_contrast_resp} Comparison of MT and V1 contrast sensitivity in humans. Figure adapted from Tootell \etal \cite{MT_contrast_Tootell}. \subref{subfig:phi_rectifier} The rectification function $\Phi$ as a function of contrast, for parameter values $\Gamma = 0.001, \rho =0.02$. This rectifier ensures high contrast sensitivity in the dorsal stream.}
\end{figure}

Various studies of the Magnocellular pathway (Shapley \etal \cite{MT_contrast_Magno_shapley1981}, Kaplan and Shapley \cite{MT_contrast_Magno_Kaplan2755}), and MT recordings of macaque (Sclar \etal \cite{MT_contrast_Sclar1990}) and humans (Tootell et al. \cite{MT_contrast_Tootell}), suggested that these neurons are highly sensitive to contrast, whereas this sensitivity in V1 neurons is much lower. Tootell \etal \cite{MT_contrast_Tootell} observed that reliable responses in MT were obtained at low contrasts, and that the responses were saturated at high contrasts (higher than $1.6\%$). Figure \ref{subfig:MT_contrast_resp}, adapted from Tootell \etal \cite{MT_contrast_Tootell}, compares the responses of V1 and MT neurons to gratings of various contrast.
Following these observations, we employ the rectification function:
\begin{equation}
\Phi = \frac{1 - e^{-R/\rho}}{1 + 1 / \Gamma e^{-R/\rho}},
\end{equation}
for dorsal neurons, where $R$ indicates cell response. Figure \ref{subfig:phi_rectifier} depicts this rectifier for $\Gamma=0.001, \rho=0.02$.

\vspace{15pt}
\textbf{Border ownership cells.} Border ownership neurons receive a feedforward signal from complex cells and a modulatory signal from MT cells. The initial border ownership signal is determined by:
\begin{equation}
B_{\theta \pm \frac{\pi}{2}}(x, y) = C_\theta(x, y) \cdot [\sum_{d_1, d_2}w(d_1, d_2)(\sum_\phi \text{MT\_ON}(x, y, \phi, d_1) +  \sum_\phi \text{MT\_OFF}(x, y, \phi, d_2))] 
\label{eq:BOS_neg}
\end{equation}
where $x, y$ specify the receptive field center. Complex cell responses, $C_\theta$, are multiplicatively modulated by on-center and off-center MT cells, represented as MT\_ON and MT\_OFF respectively, selective to orientation $\phi$. The responses of orientation-selective MT cells are summed to account for all the possible orientations, $\phi$, of the figure contour segment opposite to the occlusion boundary. In addition, $d_1, d_2$ parameters determine the center of receptive fields for the MT cells with respect to that of the border ownership neuron, and $w(\cdot, \cdot)$ is a linear weighting function with a negative slope, assigning larger weights for MT cells with receptive fields closer to the occlusion boundary. Note that each modulation term obtained by expanding the outer summation in Equation \ref{eq:BOS_neg} ($\sum_{d_1, d_2}$) is indeed a manifestation of the isotropic inhibition proposed by Shi \etal \cite{Shi_BMVC13} since each term is a weighted multiplicative process with weights set to 1 for all orientations. The outer sum, here, integrates isotropic modulatory responses at multiple visual field locations.

Now the question is how large is the surround for BOS cells? In other words, what should be the extent of region from which MT neurons provide context to the BOS cells (\ie, parameters $d_1$ and $d_2$ in Equation \ref{eq:BOS_neg})? Studying the extent of surround for V1 and V2 neurons, Shushruth \etal~\cite{shushruth2009surround} discovered that the far surround could exceed 12.5 degrees, on average about 5.5 degrees for V1 and 9.2 degrees for V2. In a similar study, Angelucci and Bullier~\cite{angelucci2003reachingBeyond} observed a similar extent of surround for neurons in V1. They tested the effect of feedback from both V2 and MT and observed that surround size could get as big as 13 times the size of V1 receptive fields, and concluded that MT neurons could provide these long-distance interactions. Accordingly, we set the extent of surround for border ownership neurons provided by MT cells to a maximum of 9$^\circ$, 13 times the average model V1 receptive field size.

\subsection*{Relaxation Labeling}
After computing the initial border ownership responses, a few iterations of relaxation labeling can provide local context and ensure smoothness in responses \cite{relaxation_zucker}. In this step, the set of border ownership neurons with receptive fields corresponding to a single visual field location comprise the set of labels associated with that location. The neighboring relationship defined over the visual field locations determines the local region that provides context to border ownership cells. Figure \ref{fig:RL_connections} illustrates the neighboring relationship in our implementation. An important component of relaxation labeling is the compatibility function between the set of labels, which determines the influence of context on a label at a visual field location and also affects the strength of this influence. 
In our implementation, a Gaussian function has been employed for compatible labels to highly reward matching labels, with a sharp fall as the labels become less similar. The penalty for incompatible labels, such as those with the same local feature selectivities but opposite side-of-figure preferences, was determined using a linear function. Relaxation labeling is restricted to 10 iterations unless convergence or a stable state is reached earlier to limit the extent of lateral propagation. Limiting the number of iterations is due to the findings of Sugihara \cite{BOwn_Sugihara2011}, who found that lateral propagation is not fast enough to travel from one side of a large figure all the way to its opposite side. That is, the impact is localized and not global. Moreover, the choice of 10 iterations for relaxation labeling was made based on the time course of border ownership processing with respect to the onset of responses reported by Zhou \etal \cite{BOwn_Zhou}.

During the iterations of relaxation labeling, only neurons with some initial responses to at least one of the labels are considered qualified for potential updates. In other words, the cells showing no initial responses to any of the labels are excluded from potential updates and neither receive nor send signals from and to their neighbors. These are specifically the cells on homogeneous regions and far from interesting features, and hence, with no initial signal from their ancestor complex cells. After the final iteration, the border ownership responses are updated as below:
\begin{equation}
R_{\text{post\_RL}}(x, y, \theta, \beta) = (1 + P(x, y, \theta, \beta)) \times R_{\text{pre\_RL}}(x, y, \theta, \beta),
\end{equation}
where $R_{\text{pre\_RL}}, R_{\text{post\_RL}}$ are responses before and after relaxation labeling, $P$ is for potentials in the range $[-0.5, 0.5]$, $\theta, \beta$ are orientation and ownership direction of the neuron, and $x, y$ signify the receptive field center. 

\section{Simulation Results}
\label{sec:results}
\begin{figure}[t!]
	\centering
	\subfigure[][]{\label{subfig:b_selectivity}\includegraphics[width=0.05\textwidth]{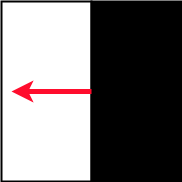}}\hspace{200pt} ~
	\subfigure[][]{\label{subfig:e_selectivity}\includegraphics[width=0.05\textwidth]{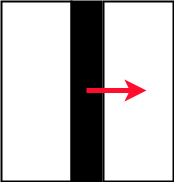}}\\
	\subfigure[]{\includegraphics[width=0.49\textwidth]{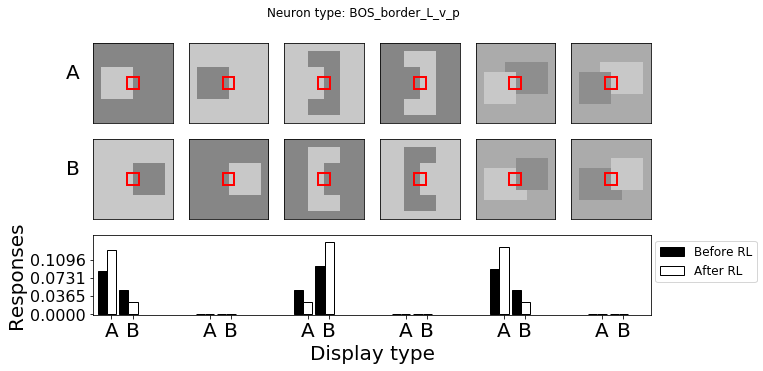}\label{subfig:border_L_v_p_resp}} ~
	\subfigure[]{\includegraphics[width=0.49\textwidth]{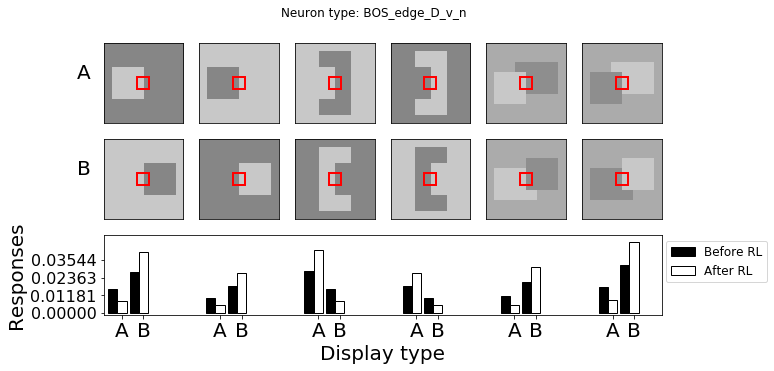}\label{subfig:edge_D_v_n_resp}}\\
	\subfigure[]{\includegraphics[width=0.49\textwidth]{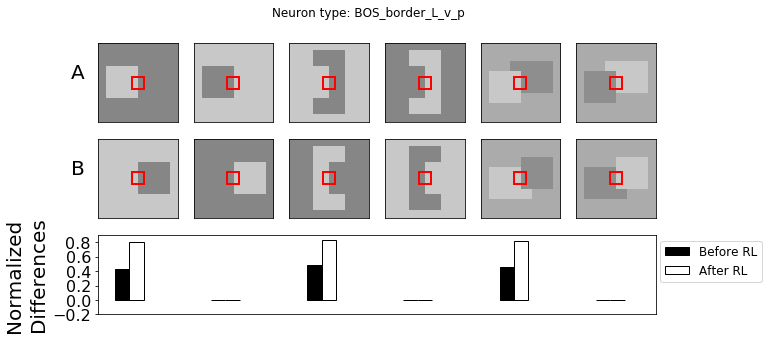}\label{subfig:norm_diff_border}} ~
	\subfigure[]{\includegraphics[width=0.49\textwidth]{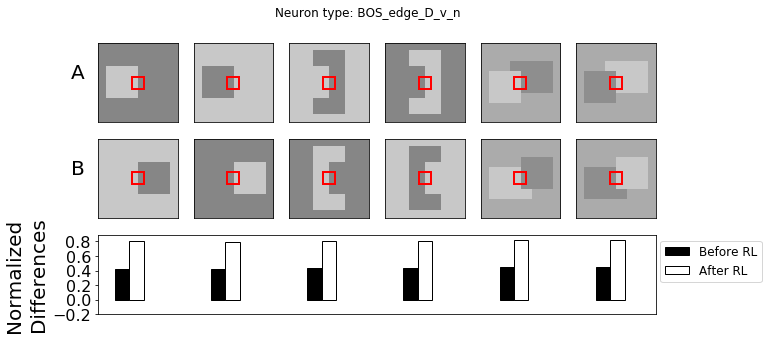}\label{subfig:norm_diff_edge}}\\
	\subfigure[]{\includegraphics[width=0.45\textwidth]{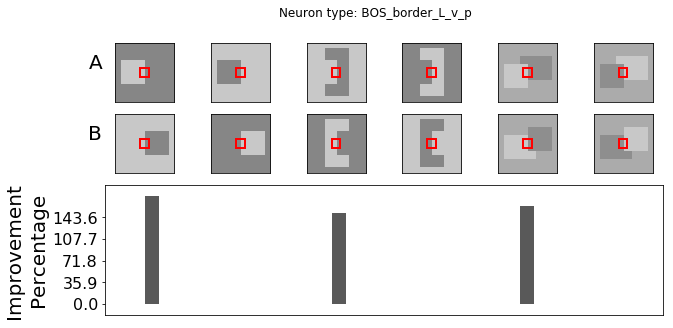}\label{subfig:imp_perc_border}} \hspace{20pt}~
	\subfigure[]{\includegraphics[width=0.45\textwidth]{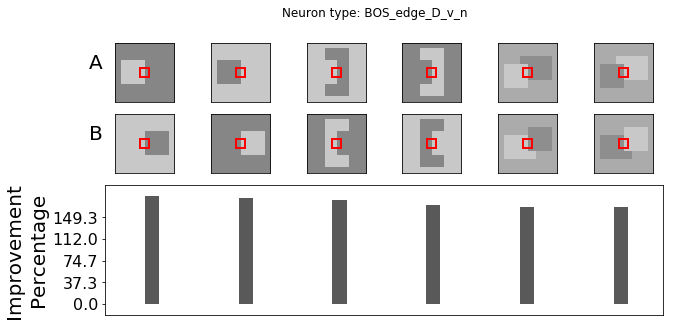}\label{subfig:imp_perc_edge}}\\
	\caption{Example of model border ownership responses to stimuli used in the neurophysiological experiments by Zhou \etal \cite{BOwn_Zhou}. \subref{subfig:b_selectivity} and \subref{subfig:e_selectivity} represent the selectivity of neurons for which responses are shown in columns of this figure. The red arrows show the border ownership direction selectivity. In \subref{subfig:border_L_v_p_resp} and \subref{subfig:edge_D_v_n_resp}, dark bars indicate the initial border ownership responses obtained by MT modulations, and the light bars show the responses after relaxation labeling. Compare the responses in \subref{subfig:border_L_v_p_resp} with the responses of a biological neurons shown in Figure \ref{subfig:BOS_bio_resp2}. In \subref{subfig:norm_diff_border} and \subref{subfig:norm_diff_edge}, the normalized differences, \ie, the relative amount of difference in responses to preferred and non-preferred stimuli are depicted. In \subref{subfig:imp_perc_border} and \subref{subfig:imp_perc_edge}, the percentage of improvement after relaxation labeling is shown.}
	\label{fig:BOS_model_responses}
\end{figure}

We tested our model on white-gray-black stimuli with intensities and shapes similar to those of the study by Zhou \etal \cite{BOwn_Zhou}. Figure \ref{subfig:border_L_v_p_resp} and \ref{subfig:edge_D_v_n_resp} depict example responses of our model border ownership neurons to these stimuli.
Figures \ref{subfig:b_selectivity} and \ref{subfig:e_selectivity} show the local feature selectivities with the red arrow indicating side-of-figure preferences. For example, the left column of plots is for a type 3 model border ownership neuron selective to vertical borders between light and dark regions on left and right of the border respectively and selective to when the figure resides on the left of the border. In Figure \ref{subfig:border_L_v_p_resp} and \ref{subfig:edge_D_v_n_resp}, the dark bars present the initial border ownership responses obtained by MT modulations. The light bars are responses after relaxation labeling. The difference of responses in these plots is clear. We encourage the reader to compare the responses in Figure \ref{subfig:border_L_v_p_resp} with the responses of a biological neuron with similar selectivity shown in Figure \ref{subfig:BOS_bio_resp2}. It is evident from these plots that the lateral modulations, \ie, relaxation labeling steps, enhance the difference of responses. In fact, in these plots, relaxation labeling improves the responses to preferred stimuli, while decreases the responses to non-preferred ones. 

In order to quantitatively measure the effect of relaxation labeling, we computed a normalized difference of responses as
\begin{equation}
D = \frac{R_\text{preferred} - R_\text{non\_preferred}}{\max(R_\text{preferred}, R_\text{non\_preferred})}.
\end{equation}
The normalized difference $D$, a value in the range $[-1, 1]$, gives a means of comparison for the strength of BOS signal differences before and after relaxation labeling. As expected, the normalized differences demonstrated in Figure \ref{subfig:norm_diff_border} and \ref{subfig:norm_diff_edge}, are larger after relaxation labeling compared to MT-modulated responses. To measure the amount of enhancement by lateral modulations, improvement percentage for before vs. after relaxation labeling was computed. As can be seen in Figure \ref{subfig:imp_perc_border} and \ref{subfig:imp_perc_edge}, more than 100\% improvement over the initial difference of responses is observed. This pattern of improvement was also seen in our other model border ownership neurons, which are not included here for brevity. The interested reader can find those responses in Appendix \ref{appendix:physio_responses}.

\begin{figure}[t!]
	\centering
	\subfigure[][]{\label{subfig:position_selectivity}\includegraphics[width=0.05\textwidth]{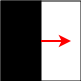}}\hspace{170pt}~~~~\\
	\subfigure[][]{\label{subfig:position_model}\includegraphics[width=0.39\textwidth]{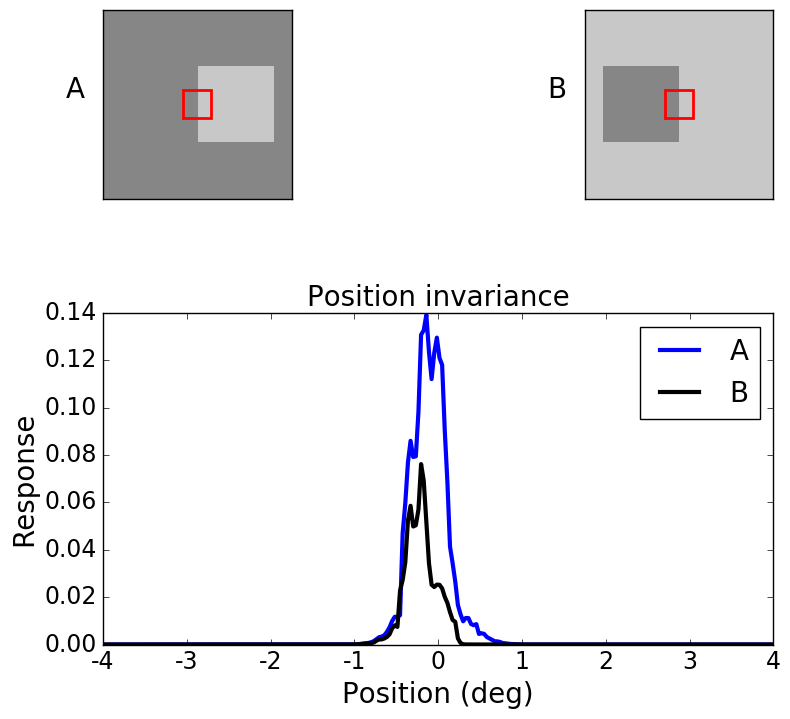}}\hspace{10pt}~
	\subfigure[][]{\label{subfig:position_bio}\includegraphics[width=0.39\textwidth]{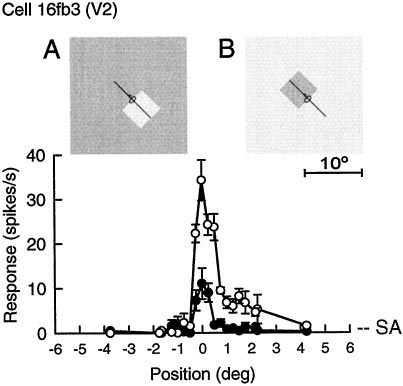}}
	\caption{Position invariance in border ownership responses. \subref{subfig:position_selectivity} Selectivity of a model border ownership cell which position invariance in its responses are shown in \subref{subfig:position_model}. \subref{subfig:position_bio} Position invariance in biological border ownership neuron responses (adapted from \cite{BOwn_Zhou}). In \subref{subfig:position_model} and \subref{subfig:position_bio}, the side of figure is aligned to the orientation selectivity of the neuron.}
	\label{fig:position_invariance}
\end{figure}

\begin{figure}[t!]
 	\centering
 	\subfigure[][]{\label{subfig:size_selectivity}\includegraphics[width=0.05\textwidth]{border_D_v_n_RF.png}}\hspace{152pt}~~~~\\
 	\subfigure[][]{\label{subfig:size_model}\includegraphics[width=0.53\textwidth]{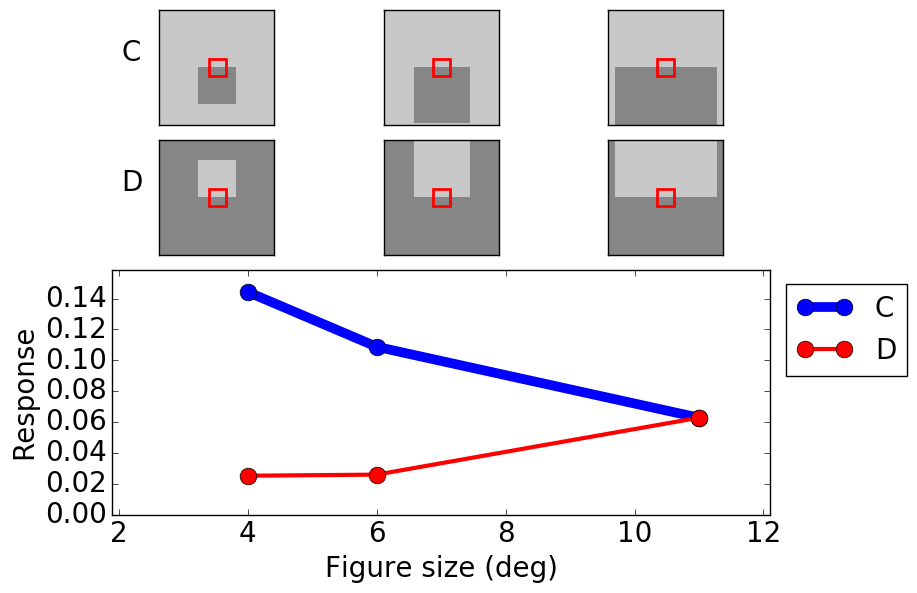}}\hspace{10pt}~
 	\subfigure[][]{\label{subfig:size_bio}\includegraphics[width=0.29\textwidth]{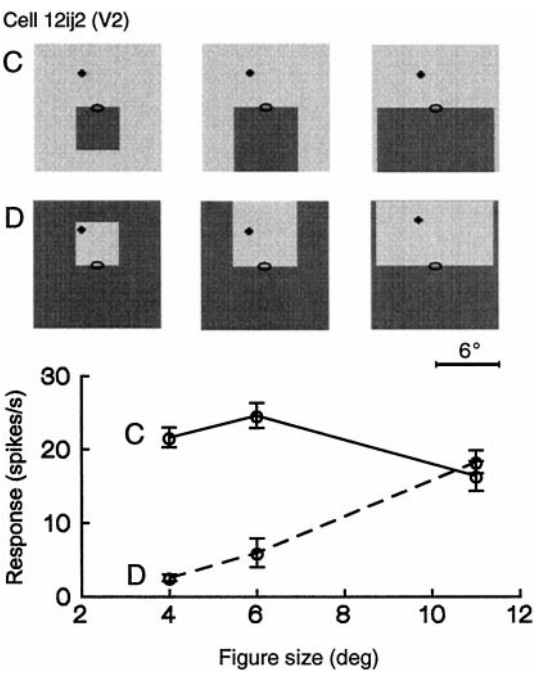}}
 	\caption{Size invariance in border ownership responses. \subref{subfig:size_selectivity} Selectivity of an example model border ownership cell with responses shown in \subref{subfig:size_model}. Our model border ownership neurons showed the difference of responses to figures of increasing sizes, up to the point that figure and ground become indistinguishable. \subref{subfig:size_bio} Responses of a biological border ownership cell to figures of various size (adapted from \cite{BOwn_Zhou}). Responses to 4$^\circ$, 6$^\circ$ and 11$^\circ$ squares are shown.}
 	\label{fig:size_invariance}
\end{figure}

\begin{figure}[t!]
	\centering
	\subfigure[][]{\label{subfig:solid_selectivity}\includegraphics[width=0.05\textwidth]{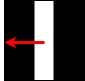}}\hspace{152pt}~~~~\\
	\subfigure[][]{\label{subfig:solid_model}\includegraphics[width=0.43\textwidth]{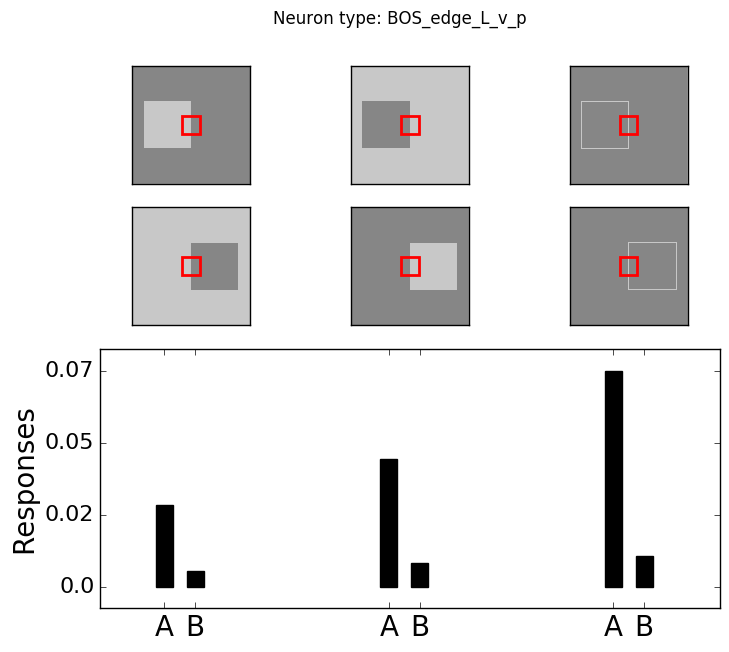}} \hspace{20pt}~
	\subfigure[][]{\label{subfig:solid_bio}\includegraphics[width=0.33\textwidth]{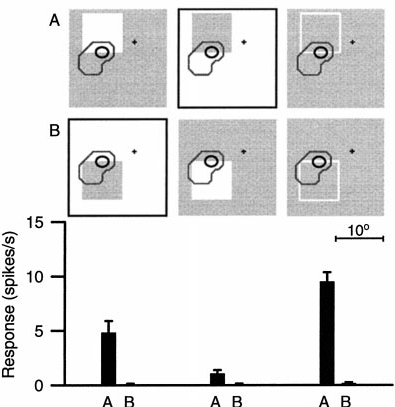}}
	\caption{Invariance to solid and outlined figures. \subref{subfig:solid_selectivity} Selectivity of an example model border ownership cell with responses to solid and outlined figures depicted in \subref{subfig:solid_model}. \subref{subfig:solid_bio} Responses of a biological border ownership neuron to solid and outlined shapes (adapted from \cite{BOwn_Zhou}).}
	\label{fig:solid_outline}
\end{figure}

Zhou \etal \cite{BOwn_Zhou} found position invariance in responses when they moved the figure perpendicular to the border. Similarly, we tested the position invariance property in our model neurons. Figure \ref{fig:position_invariance} depicts the responses of a model neuron to changes in position. One example from the neurophysiological experiment \cite{BOwn_Zhou} is included in this figure for comparison. The responses of our model border ownership cells to changes in position are comparable to those of the biological neurons, with a peak of responses when the border is located at the center of the receptive field, and a decrease in responses when it moves away from the center.
 In terms of sensitivity to figure size, the difference of responses was observed in the neurophysiological experiments up to a figure size that the figure and ground were indistinguishable. Our model neurons exhibited a similar behavior to square figures of various sizes, as shown in Figure \ref{fig:size_invariance}.
 Likewise, our model neurons demonstrated the difference of responses to both solid and outlined figures, as depicted in Figure \ref{fig:solid_outline}.

\begin{figure}[t!]
	\centering
	\subfigure[][]{\label{subfig:over_selectivity1}\includegraphics[width=0.05\textwidth]{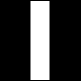}}\hspace{20pt}~
	\subfigure[][]{\label{subfig:over_selectivity2}\includegraphics[width=0.05\textwidth]{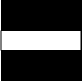}}\hspace{20pt}~\\
	\subfigure[][]{\label{subfig:over_stim1}\includegraphics[width=0.19\textwidth]{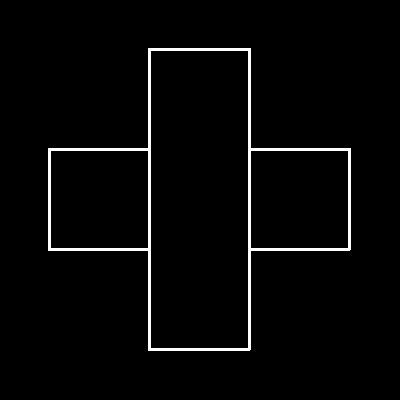}} ~
	\subfigure[][]{\label{subfig:over_vmi1}\includegraphics[width=0.23\textwidth]{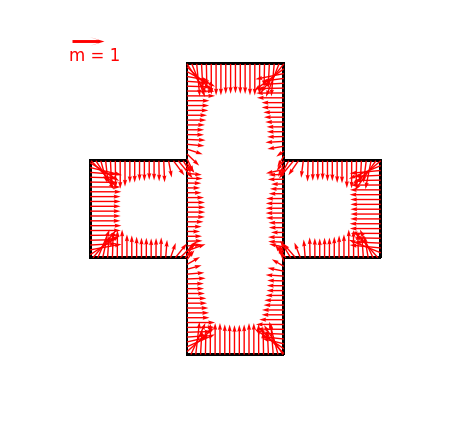}} ~
	\subfigure[][]{\label{subfig:over_dir1}\includegraphics[width=0.26\textwidth]{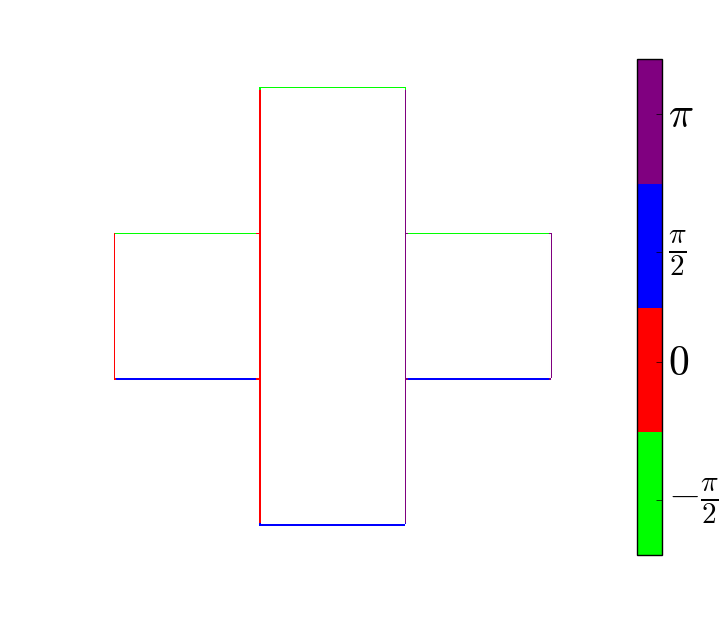}} ~
	\subfigure[][]{\label{subfig:over_activ1}\includegraphics[width=0.26\textwidth]{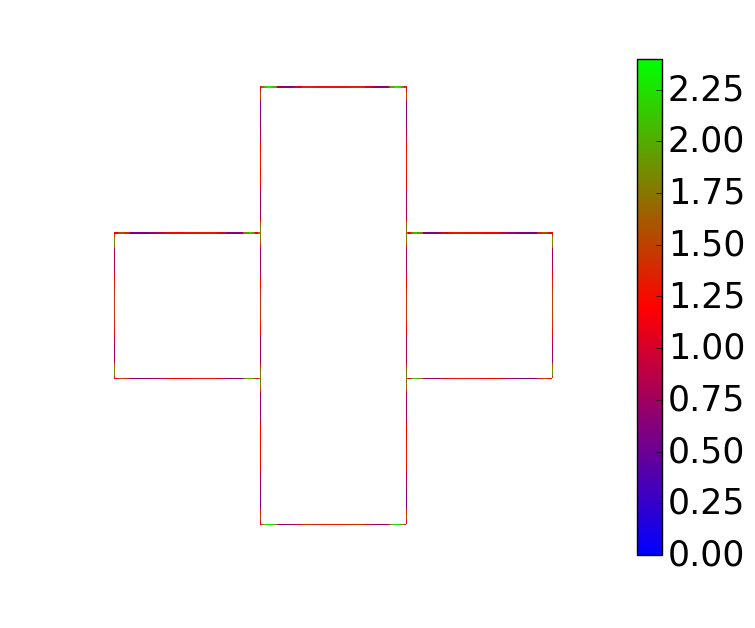}} \\
	\subfigure[][]{\label{subfig:over_stim2}\includegraphics[width=0.19\textwidth]{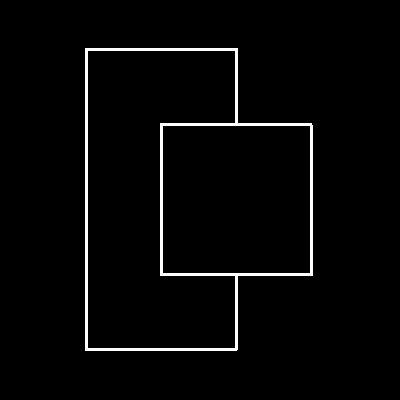}}~
	\subfigure[][]{\label{subfig:over_vmi2}\includegraphics[width=0.23\textwidth]{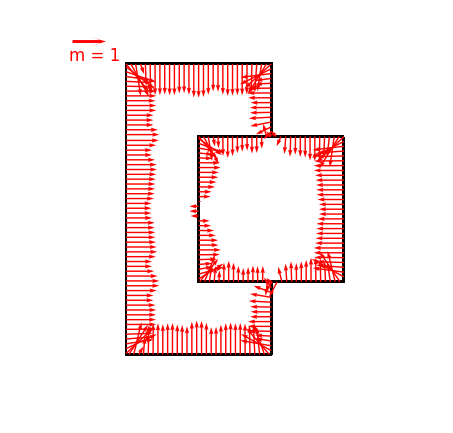}}~
	\subfigure[][]{\label{subfig:over_dir2}\includegraphics[width=0.26\textwidth]{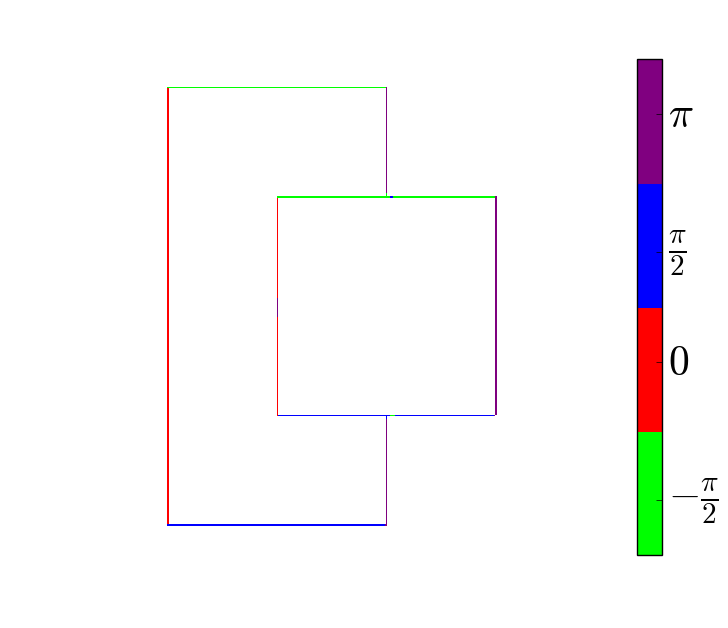}} ~
	\subfigure[][]{\label{subfig:over_activ2}\includegraphics[width=0.26\textwidth]{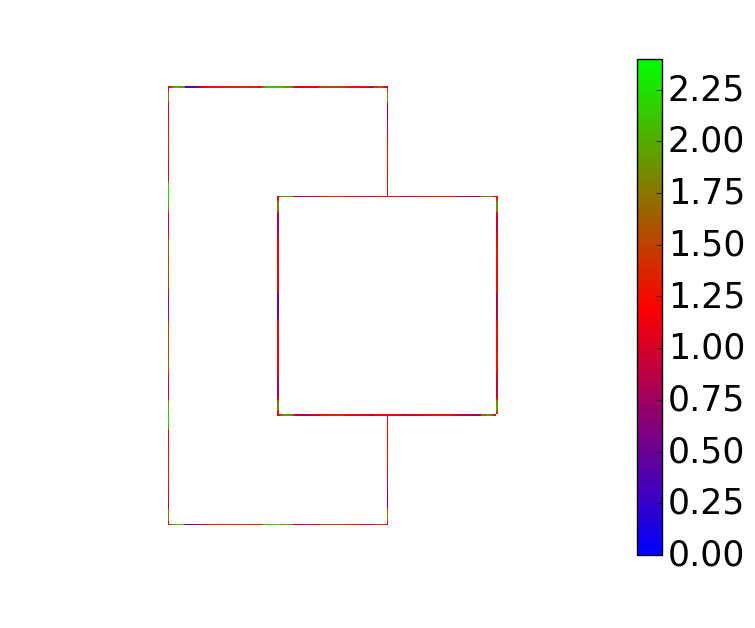}}\\
	\caption{Examples presenting the performance of our model border ownership neurons to two stimuli of overlapping shapes. The results are for the pool of cells selective to light bars on dark ground as indicated by the local features for these cells in \subref{subfig:over_selectivity1} and \subref{subfig:over_selectivity2}, over all orientations and side-of-figure preferences in the model. \subref{subfig:over_stim1} and \subref{subfig:over_stim2} show the input to the network, \subref{subfig:over_vmi1} and \subref{subfig:over_vmi2} present the vectorial modulation index \cite{BOwn_Craft} results overlayed on input images with intensities inverted, \subref{subfig:over_dir1} and \subref{subfig:over_dir2} are for the color-coded direction of responses of neurons with maximum responses among all those in the pool, along contours of figures in input, while \subref{subfig:over_activ1} and \subref{subfig:over_activ2} show the strength of these responses. }
	\label{fig:vmi_results}
\end{figure}

The length and direction of the vector obtained from the vectorial modulation index (VMI) introduced by Craft \etal \cite{BOwn_Craft} describes the strength and side-of-figure direction of the border ownership signal utilizing a vector. For a couple of examples of overlapping shapes, the performance of our model border ownership cells is demonstrated in Figure \ref{fig:vmi_results}. These examples include the responses of the pool of border ownership neurons, selective to all the implemented orientations and side-of-figure preferences with local feature selectivities shown in \ref{subfig:over_selectivity1} and \ref{subfig:over_selectivity2}. The first column in each row shows the input to the network, and the second column is VMI vectors overlayed on the input with inverted intensities. The third and fourth columns in this figure are for the direction and signal strength of the border ownership cell with maximum response among the neurons in the pool. These examples show that even though we imposed no priors over shape or T-junctions and no feedback from V4, the border ownership signals are in agreement with the perception of two overlapping shapes.

\begin{figure}[t!]
	\centering
	\subfigure[][]{\label{subfig:kan_selectivity1}\includegraphics[width=0.05\textwidth]{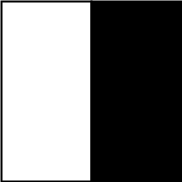}}\hspace{30pt}~
	\subfigure[][]{\label{subfig:kan_selectivity2}\includegraphics[width=0.05\textwidth]{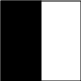}}\hspace{30pt}~
	\subfigure[][]{\label{subfig:kan_selectivity3}\includegraphics[width=0.05\textwidth]{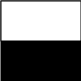}}\hspace{30pt}~
	\subfigure[][]{\label{subfig:kan_selectivity4}\includegraphics[width=0.05\textwidth]{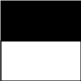}}\hspace{10pt}\\
	\subfigure[][]{\label{subfig:kan1_stim}\includegraphics[width=0.19\textwidth]{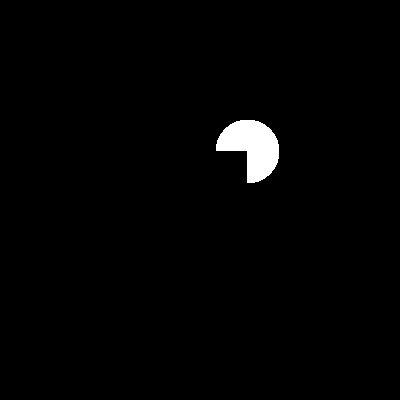}} ~
	\subfigure[][]{\label{subfig:kan1_vmi}\includegraphics[width=0.23\textwidth]{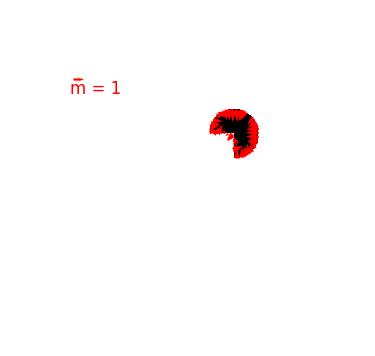}} ~
	\subfigure[][]{\label{subfig:kan1_dir}\includegraphics[width=0.26\textwidth]{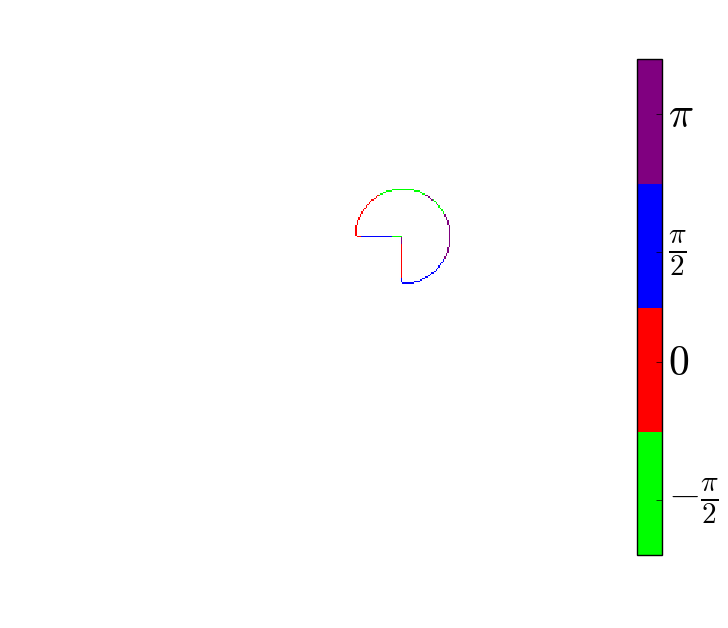}} ~
	\subfigure[][]{\label{subfig:kan1_activ}\includegraphics[width=0.26\textwidth]{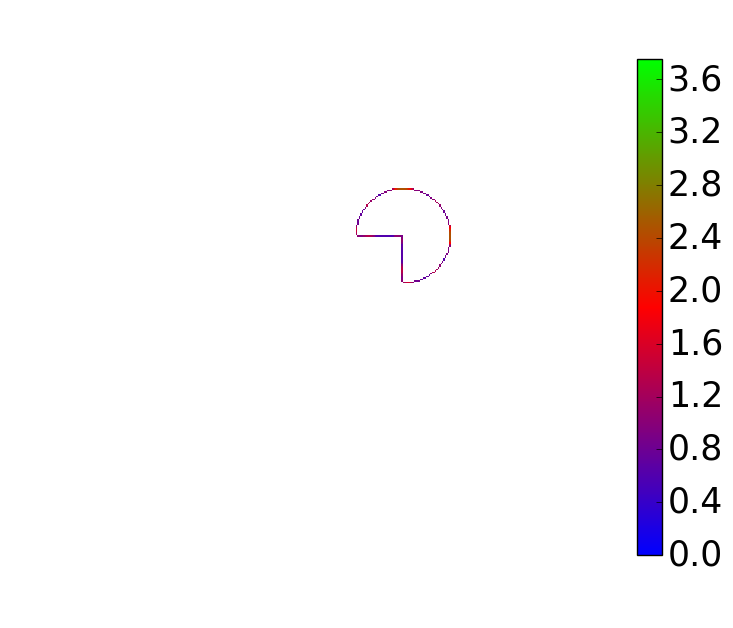}} \\
	\subfigure[][]{\label{subfig:kan2_stim}\includegraphics[width=0.19\textwidth]{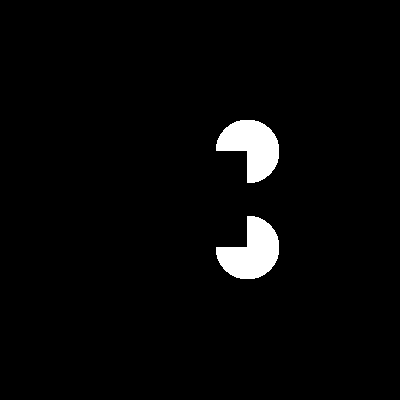}} ~
	\subfigure[][]{\label{subfig:kan2_vmi}\includegraphics[width=0.23\textwidth]{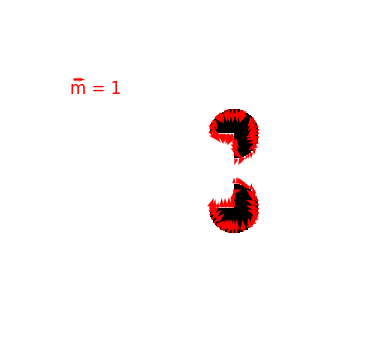}} ~
	\subfigure[][]{\label{subfig:kan2_dir}\includegraphics[width=0.26\textwidth]{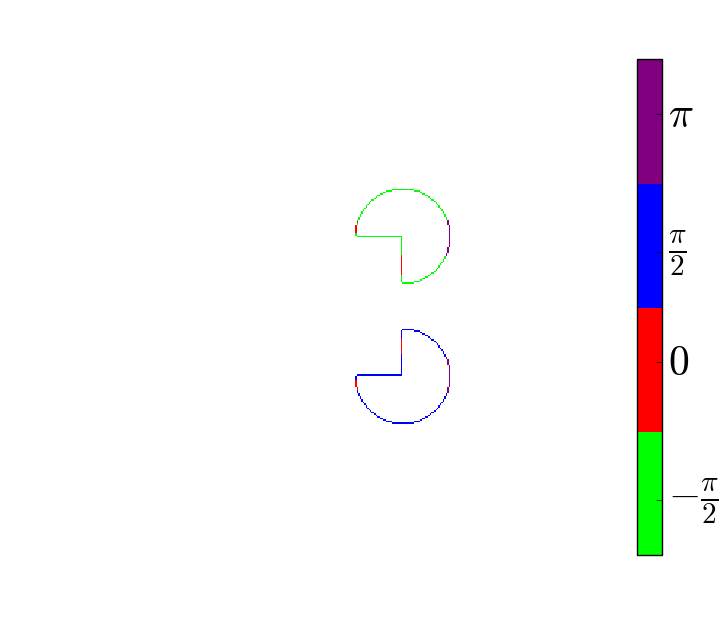}} ~
	\subfigure[][]{\label{subfig:kan2_activ}\includegraphics[width=0.26\textwidth]{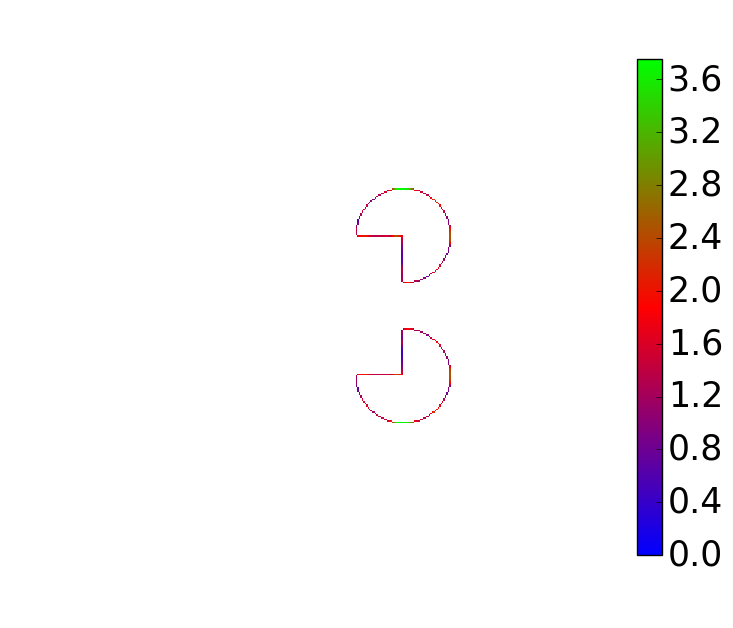}} \\
	\subfigure[][]{\label{subfig:kan4_stim}\includegraphics[width=0.19\textwidth]{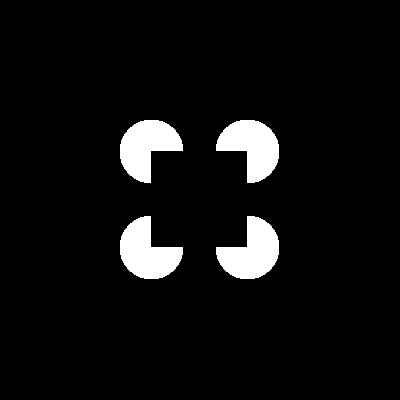}} ~
	\subfigure[][]{\label{subfig:kan4_vmi}\includegraphics[width=0.23\textwidth]{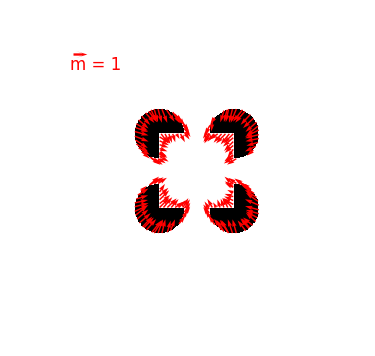}} ~
	\subfigure[][]{\label{subfig:kan4_dir}\includegraphics[width=0.26\textwidth]{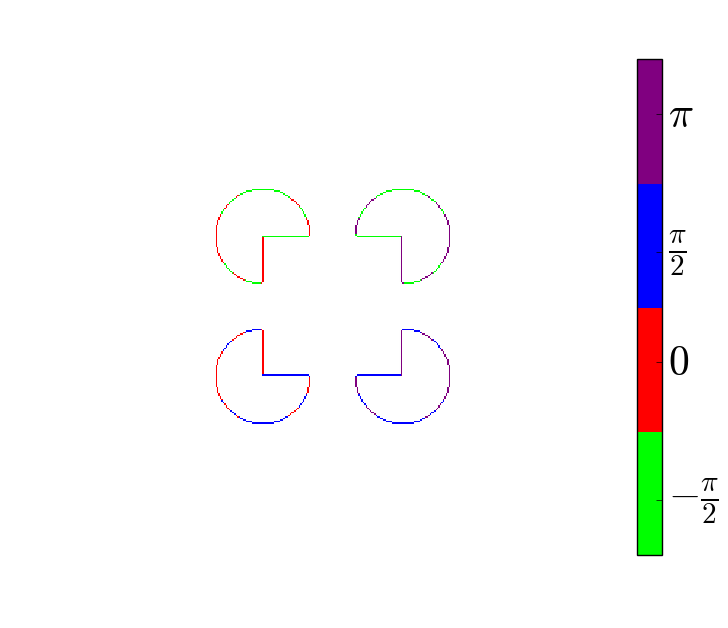}} ~
	\subfigure[][]{\label{subfig:kan4_activ}\includegraphics[width=0.26\textwidth]{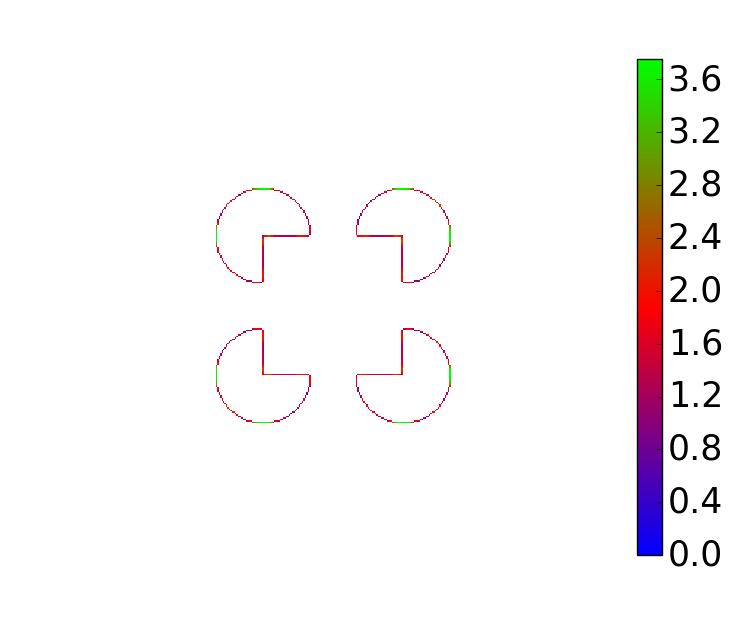}} 
	\caption{Examples presenting the performance of our model border ownership neurons to one, two and four Pac-Mans present in the stimuli. The results are for the pool of cells selective to borders between light and dark regions at horizontal and vertical orientations and all BOS direction selectivities shown in \subref{subfig:kan_selectivity1}, \subref{subfig:kan_selectivity2}, \subref{subfig:kan_selectivity3}, and \subref{subfig:kan_selectivity4}. Figures \subref{subfig:kan1_stim}, \subref{subfig:kan2_stim}, and \subref{subfig:kan4_stim} show the input to the network. The vectorial modulation index results overlayed on input images with intensities inverted are presented in \subref{subfig:kan1_vmi}, \subref{subfig:kan2_vmi}, and \subref{subfig:kan4_vmi}. In \subref{subfig:kan1_dir}, \subref{subfig:kan2_dir}, and \subref{subfig:kan4_dir}, the direction of responses of neurons with maximum responses among all those in the pool, along contours of figures in input, are depicted. The directions in these figures are color-coded. Finally, \subref{subfig:kan1_activ}, \subref{subfig:kan2_activ}, and \subref{subfig:kan4_activ} present the strength of these responses. In the example of a single Pac-Man, the direction of BOS responses with maximum response shown in \subref{subfig:kan1_dir} clearly shows that the border ownership direction for the the two edge components of the concavity are correctly assigned except for the very corner of the concavity. In other words, the border ownership cells see an isolated figure in stimuli. However, the direction of this assignment gradually changes when two and later four Pac-Mans are present in the display, as depicted in \subref{subfig:kan2_dir} and \subref{subfig:kan4_dir}.}
	\label{fig:Kanizsa_vmi_results}
\end{figure}

Figure \ref{fig:Kanizsa_vmi_results} includes examples of one, two and four Pac-Mans gradually forming a Kanizsa's square. The vectorial modulation index vectors and the direction of neurons with maximum responses show that in the presence of a single Pac-Man, the sole figure is perceived as a single figure. However, when more Pac-Mans are added, the border ownership signal is shifting the perception from isolated Pac-Mans to two white circles occluded by a black bar on one side in case of the two Pac-Mans, and a dark square that occluded four white circles for the stimulus with four Pac-Mans.

\section{Conclusion}
\label{sec:conclusion}
Border ownership encoding happens early in the ventral pathway, suggesting border ownership assignment being a step toward object recognition. Border ownership assignment depends not only on the local features but also on contextual information. The time course of the border ownership signal on figures of various sizes suggests that feedforward signals with lateral modulations cannot provide contextual information determining the ownership at occlusion boundaries, but feedback from higher areas is required. Moreover, the time course of border ownership cells and those of IT neurons indicate that feedback from IT is not fast enough to provide context for BOS cells. Likewise, here, we described neurophysiological evidence suggesting that V4 is not fast enough in terms of both linear and nonlinear processing components to send a timely feedback signal to border ownership cells.

Inspired by these observations, we introduced a hierarchical model that not only replicates the behavior of biological BOS cells but, in contrast to previous models, is also biologically plausible and could describe the mechanism the brain employs for border ownership assignment. This was achieved through early recurrence from the dorsal stream, which could very well describe the observed difference of responses in biological border ownership cells from the beginning of stimulus onset. Aside from fast latencies in the dorsal stream, neurons in this pathway are highly sensitive to spatiotemporal variations at coarser scales compared to those of the ventral stream and are less sensitive to contrast. As a result, the signal from the dorsal stream becomes more robust to noise and low contrasts at occlusion boundaries. These two characteristics of MT cells and their short latencies make them perfect candidates for providing contextual information to border ownership cells. Previously, early recurrence from the dorsal stream showed improvements in low-level feature representations \cite{Shi2012biologically} and edge detection \cite{Shi_BMVC13}. In a similar attempt, we demonstrated the role of early recurrence from the dorsal stream in providing global information essential for even higher levels of abstraction, \ie, border ownership assignment.

Another novel aspect of our model is combining global and local context. Although the global information provided by MT cells results in a border ownership assignment, it does not warrant that the border ownership directions are consistent among neighboring visual field locations. The relaxation labeling step in our algorithm implementing lateral connections among border ownership cells ensures consistent labeling in local regions. Our results clearly demonstrate an enhancement in the border ownership signal after a few iterations of relaxation labeling. The improvement in response differences to preferred and non-preferred stimuli was more than 100\% after relaxation labeling. Additionally, this step could describe the gradual increase in the difference of responses as the iterations of relaxation labeling provide local support with a lag to border ownership neurons, due to the time required for the signal to travel in a local neighborhood. 

In our simulation results, we demonstrated that our model neurons of types 1 and 3 show side-of-figure preferences similar to those of biological cells, and are invariant to position, size and solid/outlined figures. Furthermore, when presented with overlapping shapes, our model BOS cells assign borders to the occluding figure along the occlusion boundaries. Interestingly, in the experiment with Pac-Mann figures, the borders are assigned to the Pac-Man in case of a single figure, and then to an illusory occluding shape with the gradual addition of more Pac-Mans to form the Kanizsa's square.

The current model is limited to horizontal and vertical orientations and one step to further improve this model is to implement a variety of orientations. Zhou \etal \cite{BOwn_Zhou} reported border ownership cells selective to color borders and designed their stimuli to match the selectivity of those neurons. Another future attempt is to add color borders to our model. Moreover, we would like to test the performance of our model to more complex stimuli such as randomly generated overlapping polygonal scenes and real images. A natural extension to our model would be learning the set of free parameters in our network, such as kernel parameters in each layer and the extent of surround for border ownership neurons.

As a final remark, although neurophysiological evidence regarding latencies as well as our experimental results support the hypothesis of dorsal and lateral modulations for border ownership assignment, it will be insightful to examine this hypothesis on biological cells. One possible experiment is to remove any feedback from the dorsal stream to these neurons. By removing the dorsal effect, the border ownership assignment would have to be based on feedforward and lateral connections, a hypothesis that has already been ruled out \cite{BOwn_Zhang2010, BOwn_Sugihara2011}. As a result, we expect the border ownership neurons to have no explicit representation of the ownership assignment at occlusion boundaries.
\section{Acknowledgments}
This research was supported by several sources for which the authors are grateful: Air Force Office of Scientific Research (FA9550-14-1-0393), the Canada Research Chairs Program, and the Natural Sciences and Engineering Research Council of Canada.
\bibliographystyle{ieeetr}
\bibliography{../references/references.bib}

\newpage
\begin{appendices}
\section{}
\label{appendix:physio_responses}
Here, we provide the responses of our model neurons to stimuli employed in the neurophysiological study by Zhou \etal \cite{BOwn_Zhou}, similar to those shown in Figure \ref{fig:BOS_model_responses}.
\begin{figure}[h!]
	\centering
	\subfigure[][]{\label{subfig:b2_selectivity}\includegraphics[width=0.05\textwidth]{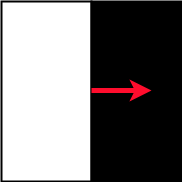}}\hspace{200pt} ~
	\subfigure[][]{\label{subfig:e2_selectivity}\includegraphics[width=0.05\textwidth]{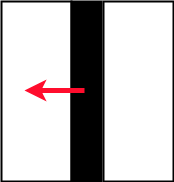}}\\
	\subfigure[]{\includegraphics[width=0.49\textwidth]{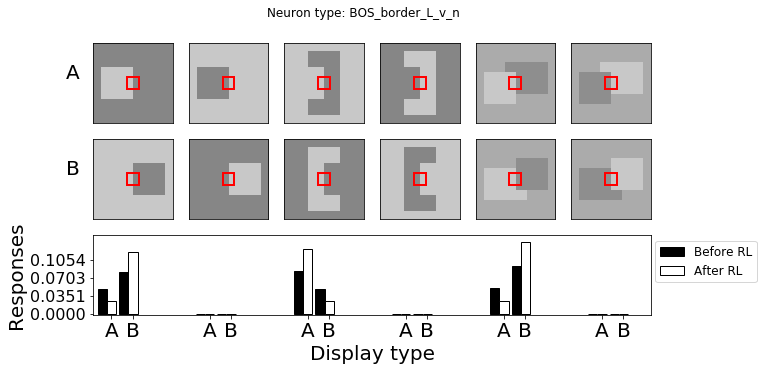}\label{subfig:border_L_v_n_resp}} ~
	\subfigure[]{\includegraphics[width=0.49\textwidth]{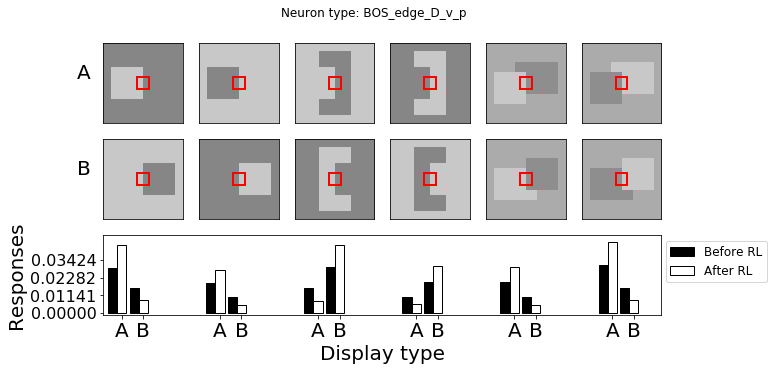}\label{subfig:edge_D_v_p_resp}}\\
	\subfigure[]{\includegraphics[width=0.49\textwidth]{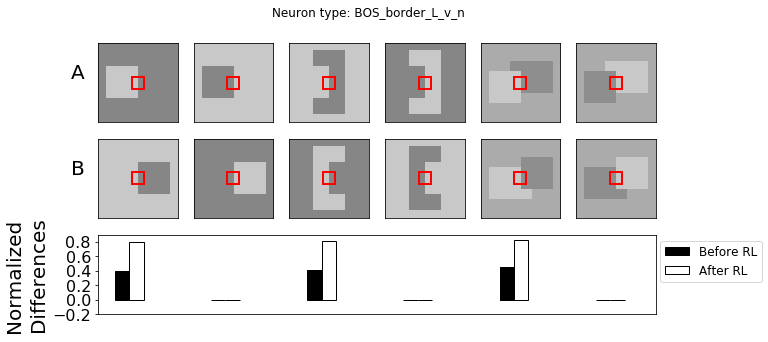}\label{subfig:b2_norm_diff_border}} ~
	\subfigure[]{\includegraphics[width=0.49\textwidth]{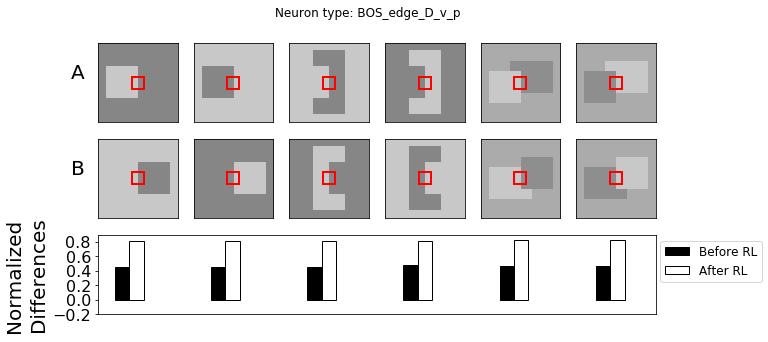}\label{subfig:e2_norm_diff_edge}}\\
	\subfigure[]{\includegraphics[width=0.45\textwidth]{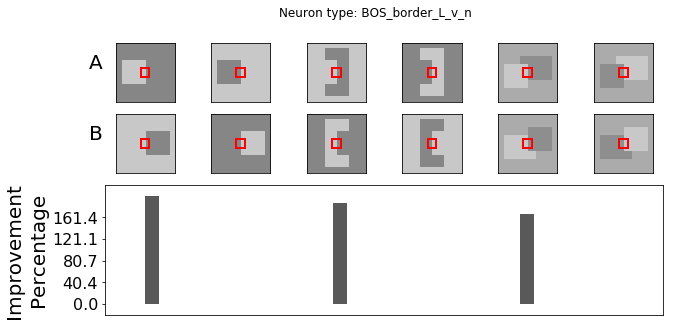}\label{subfig:b2_imp_perc_border}} \hspace{20pt}~
	\subfigure[]{\includegraphics[width=0.45\textwidth]{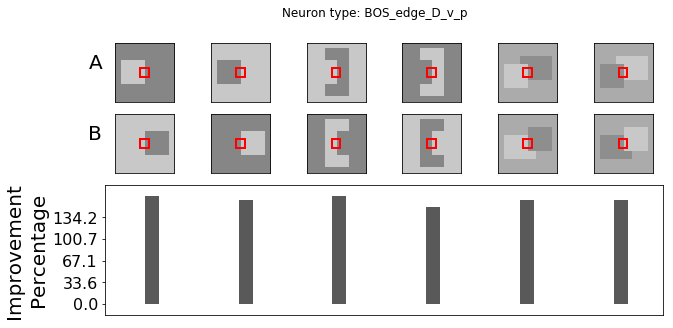}\label{subfig:e2_imp_perc_edge}}\\
	\caption{Example of model border ownership responses to stimuli used in the neurophysiological experiments by Zhou \etal \cite{BOwn_Zhou}. \subref{subfig:b2_selectivity} and \subref{subfig:e2_selectivity} represent the selectivity of neurons for which responses are shown in columns of this figure. The red arrows show the border ownership direction selectivity. In \subref{subfig:border_L_v_n_resp} and \subref{subfig:edge_D_v_p_resp}, dark bars indicate the initial border ownership responses obtained by MT modulations, and the light bars show the responses after relaxation labeling. In \subref{subfig:b2_norm_diff_border} and \subref{subfig:e2_norm_diff_edge}, the normalized differences, \ie, the relative amount of difference in responses to preferred and non-preferred stimuli are depicted. In \subref{subfig:b2_imp_perc_border} and \subref{subfig:e2_imp_perc_edge}, the percentage of improvement after relaxation labeling is shown.}
	\label{fig:BOS_model_responses2}
\end{figure}
\begin{figure}[]
	\centering
	\subfigure[][]{\label{subfig:b3_selectivity}\includegraphics[width=0.05\textwidth]{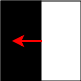}}\hspace{200pt} ~
	\subfigure[][]{\label{subfig:e3_selectivity}\includegraphics[width=0.05\textwidth]{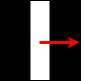}}\\
	\subfigure[]{\includegraphics[width=0.49\textwidth]{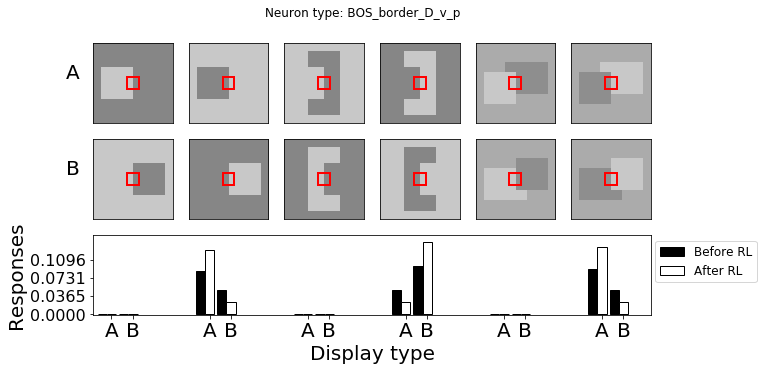}\label{subfig:border_D_v_p_resp}} ~
	\subfigure[]{\includegraphics[width=0.49\textwidth]{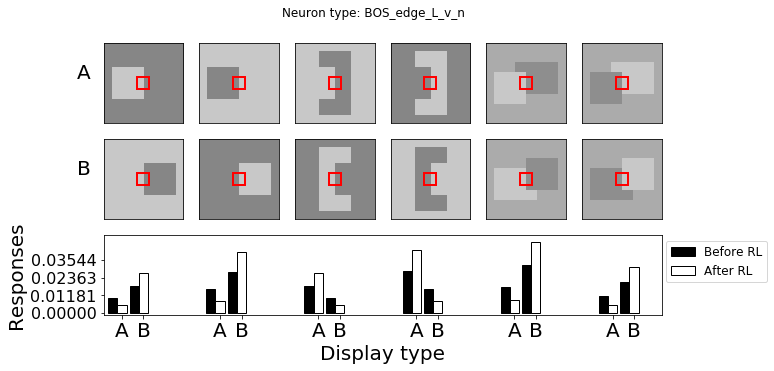}\label{subfig:edge_L_v_n_resp}}\\
	\subfigure[]{\includegraphics[width=0.49\textwidth]{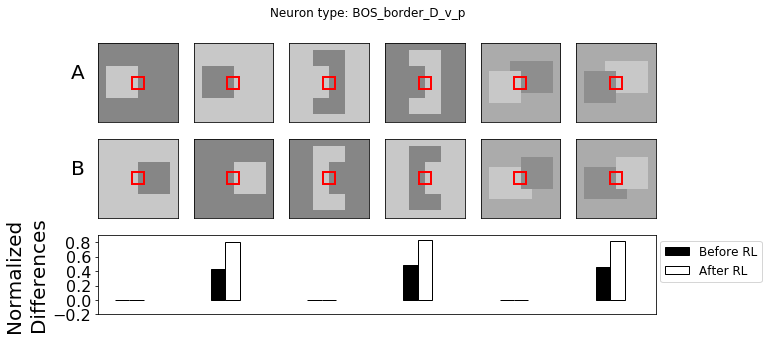}\label{subfig:b3_norm_diff_border}} ~
	\subfigure[]{\includegraphics[width=0.49\textwidth]{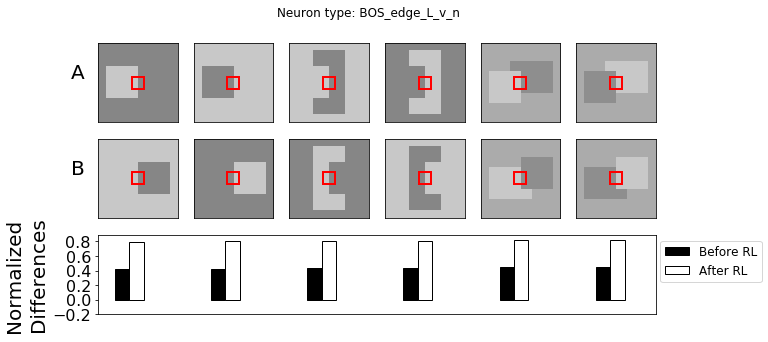}\label{subfig:e3_norm_diff_edge}}\\
	\subfigure[]{\includegraphics[width=0.45\textwidth]{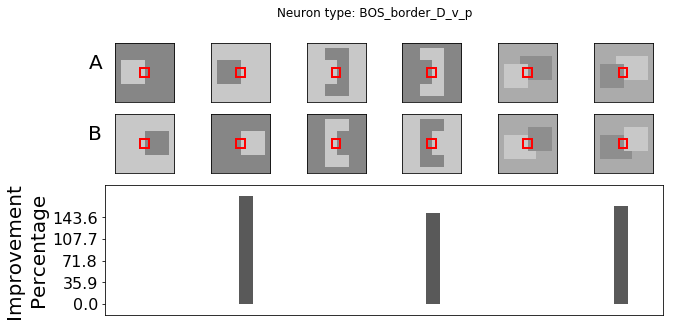}\label{subfig:b3_imp_perc_border}} \hspace{20pt}~
	\subfigure[]{\includegraphics[width=0.45\textwidth]{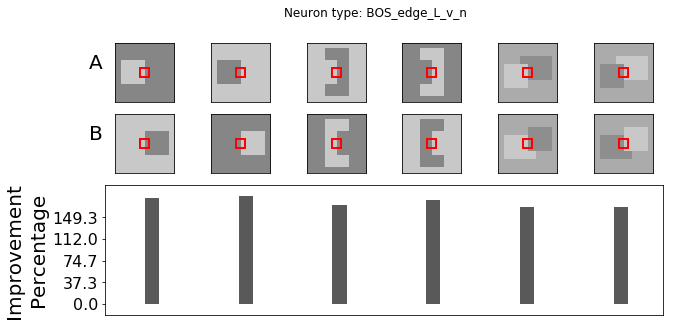}\label{subfig:e3_imp_perc_edge}}\\
	\caption{Example of model border ownership responses to stimuli used in the neurophysiological experiments by Zhou \etal \cite{BOwn_Zhou}. \subref{subfig:b3_selectivity} and \subref{subfig:e3_selectivity} represent the selectivity of neurons for which responses are shown in columns of this figure. The red arrows show the border ownership direction selectivity. In \subref{subfig:border_D_v_p_resp} and \subref{subfig:edge_L_v_n_resp}, dark bars indicate the initial border ownership responses obtained by MT modulations, and the light bars show the responses after relaxation labeling. In \subref{subfig:b3_norm_diff_border} and \subref{subfig:e3_norm_diff_edge}, the normalized differences, \ie, the relative amount of difference in responses to preferred and non-preferred stimuli are depicted. In \subref{subfig:b3_imp_perc_border} and \subref{subfig:e3_imp_perc_edge}, the percentage of improvement after relaxation labeling is shown.}
	\label{fig:BOS_model_responses3}
\end{figure}
\begin{figure}[]
	\centering
	\subfigure[][]{\label{subfig:b4_selectivity}\includegraphics[width=0.05\textwidth]{border_D_v_n_RF.png}}\hspace{200pt} ~
	\subfigure[][]{\label{subfig:e4_selectivity}\includegraphics[width=0.05\textwidth]{border_edge_L_v_p_RF.png}}\\
	\subfigure[]{\includegraphics[width=0.49\textwidth]{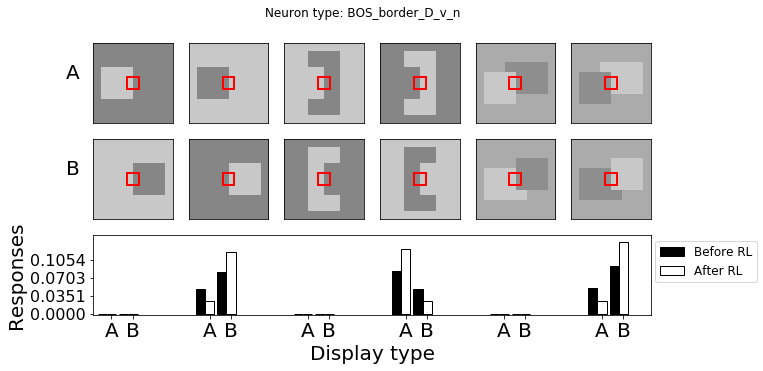}\label{subfig:border_D_v_n_resp}} ~
	\subfigure[]{\includegraphics[width=0.49\textwidth]{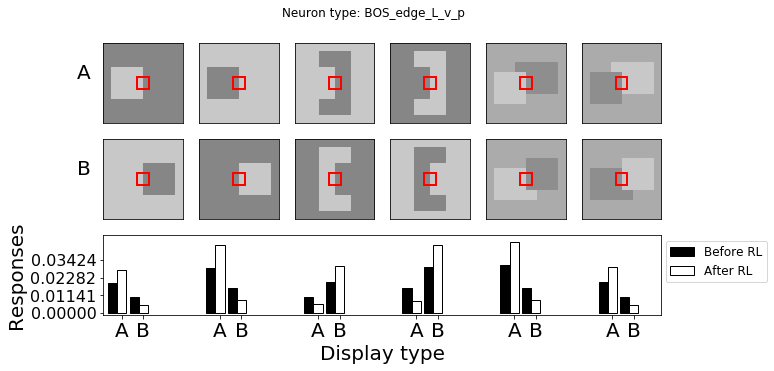}\label{subfig:edge_L_v_p_resp}}\\
	\subfigure[]{\includegraphics[width=0.49\textwidth]{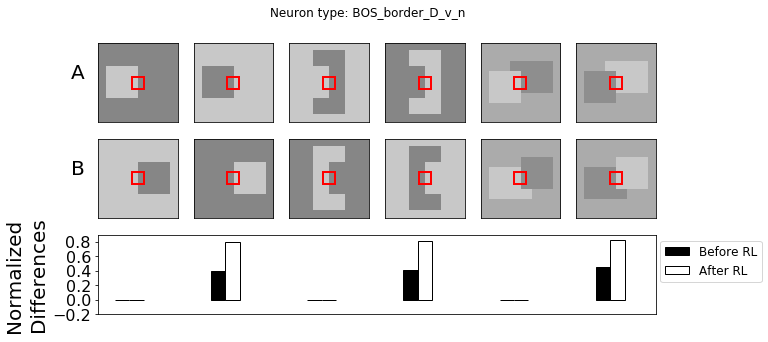}\label{subfig:b4_norm_diff_border}} ~
	\subfigure[]{\includegraphics[width=0.49\textwidth]{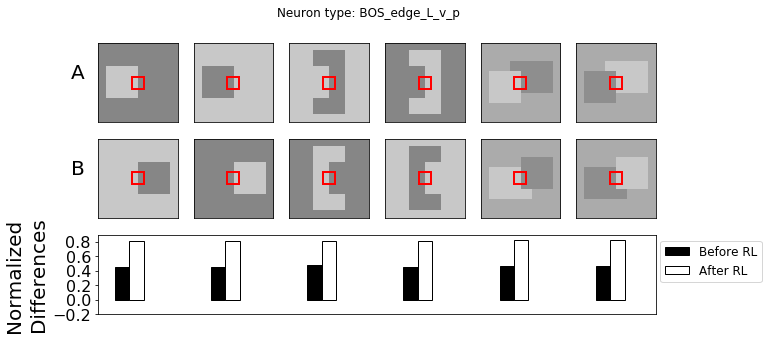}\label{subfig:e4_norm_diff_edge}}\\
	\subfigure[]{\includegraphics[width=0.45\textwidth]{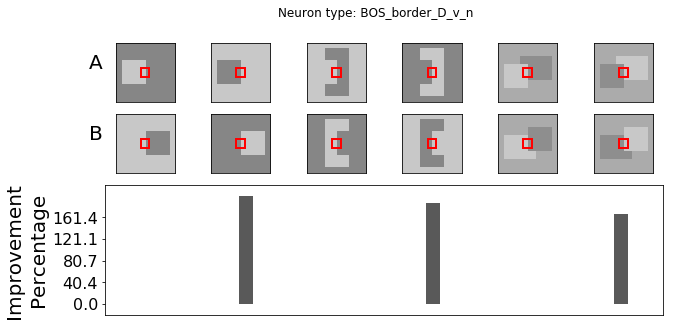}\label{subfig:b4_imp_perc_border}} \hspace{20pt}~
	\subfigure[]{\includegraphics[width=0.45\textwidth]{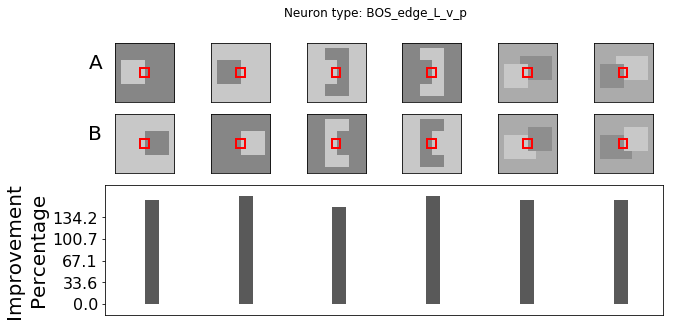}\label{subfig:e4_imp_perc_edge}}\\
	\caption{Example of model border ownership responses to stimuli used in the neurophysiological experiments by Zhou \etal \cite{BOwn_Zhou}. \subref{subfig:b4_selectivity} and \subref{subfig:e4_selectivity} represent the selectivity of neurons for which responses are shown in columns of this figure. The red arrows show the border ownership direction selectivity. In \subref{subfig:border_D_v_n_resp} and \subref{subfig:edge_L_v_p_resp}, dark bars indicate the initial border ownership responses obtained by MT modulations, and the light bars show the responses after relaxation labeling. In \subref{subfig:b4_norm_diff_border} and \subref{subfig:e4_norm_diff_edge}, the normalized differences, \ie, the relative amount of difference in responses to preferred and non-preferred stimuli are depicted. In \subref{subfig:b4_imp_perc_border} and \subref{subfig:e4_imp_perc_edge}, the percentage of improvement after relaxation labeling is shown.}
	\label{fig:BOS_model_response43}
\end{figure}
\begin{figure}[h!]
	\centering
	\subfigure[][]{\label{subfig:b5_selectivity}\includegraphics[width=0.05\textwidth]{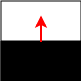}}\hspace{200pt} ~
	\subfigure[][]{\label{subfig:e5_selectivity}\includegraphics[width=0.05\textwidth]{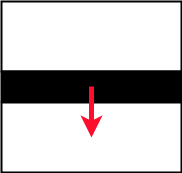}}\\
	\subfigure[]{\includegraphics[width=0.49\textwidth]{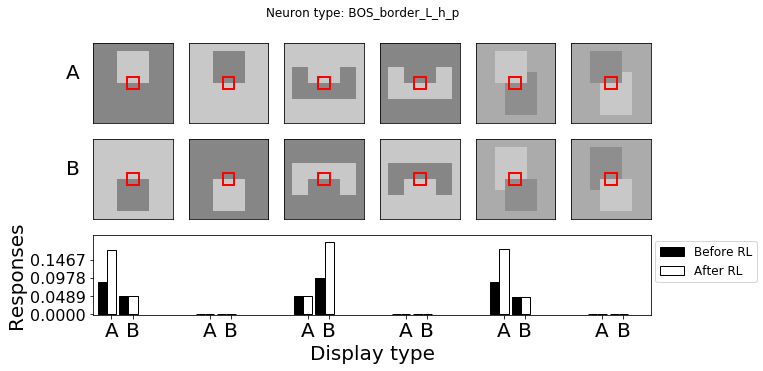}\label{subfig:border_L_h_p_resp}} ~
	\subfigure[]{\includegraphics[width=0.49\textwidth]{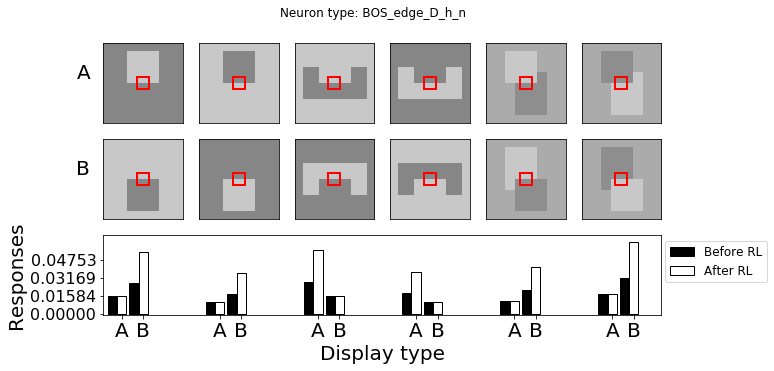}\label{subfig:edge_D_h_n_resp}}\\
	\subfigure[]{\includegraphics[width=0.49\textwidth]{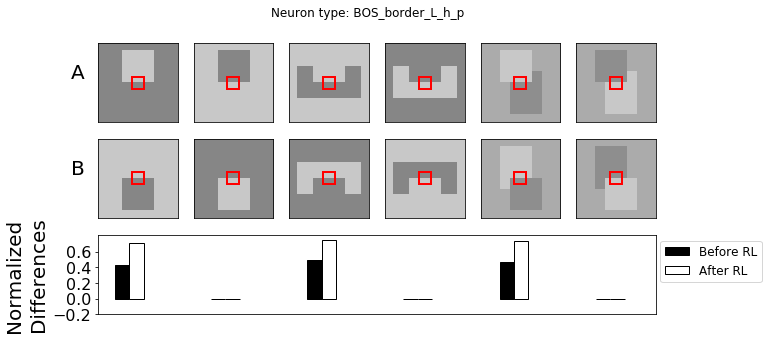}\label{subfig:b5_norm_diff_border}} ~
	\subfigure[]{\includegraphics[width=0.49\textwidth]{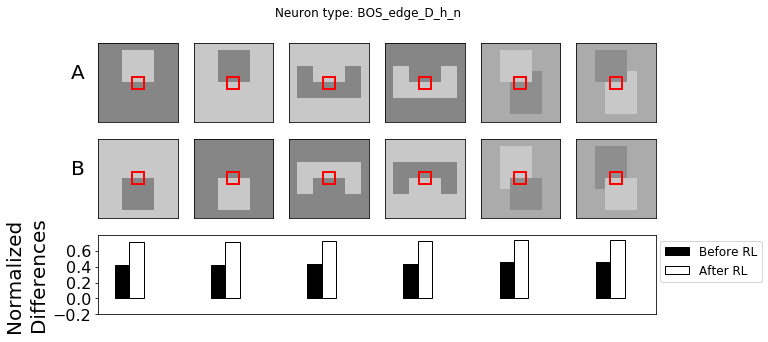}\label{subfig:e5_norm_diff_edge}}\\
	\subfigure[]{\includegraphics[width=0.45\textwidth]{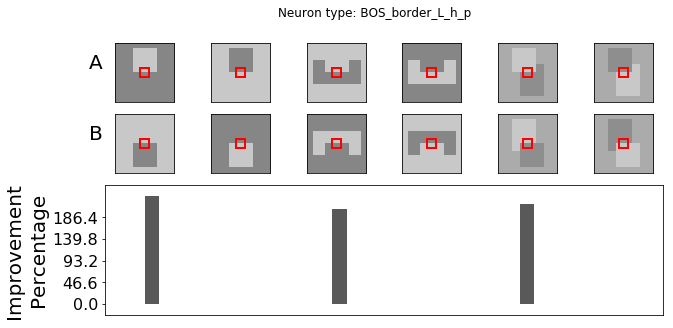}\label{subfig:b5_imp_perc_border}} \hspace{20pt}~
	\subfigure[]{\includegraphics[width=0.45\textwidth]{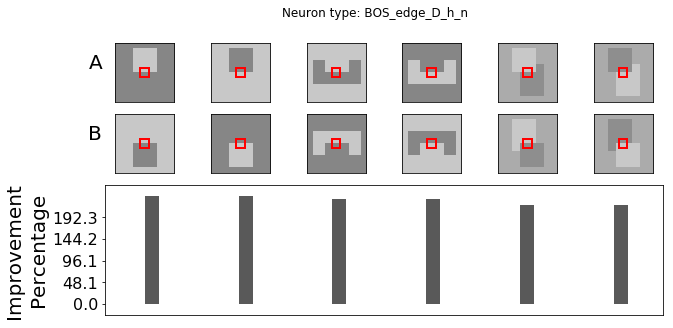}\label{subfig:e5_imp_perc_edge}}\\
	\caption{Example of model border ownership responses to stimuli used in the neurophysiological experiments by Zhou \etal \cite{BOwn_Zhou}. Note that the stimuli are rotated to match the orientation selectivity of the neuron. \subref{subfig:b5_selectivity} and \subref{subfig:e5_selectivity} represent the selectivity of neurons for which responses are shown in columns of this figure. The red arrows show the border ownership direction selectivity. In \subref{subfig:border_L_h_p_resp} and \subref{subfig:edge_D_h_n_resp}, dark bars indicate the initial border ownership responses obtained by MT modulations, and the light bars show the responses after relaxation labeling. In \subref{subfig:b5_norm_diff_border} and \subref{subfig:e5_norm_diff_edge}, the normalized differences, \ie, the relative amount of difference in responses to preferred and non-preferred stimuli are depicted. In \subref{subfig:b5_imp_perc_border} and \subref{subfig:e5_imp_perc_edge}, the percentage of improvement after relaxation labeling is shown.}
	\label{fig:BOS_model_responses5}
\end{figure}
\begin{figure}[h!]
	\centering
	\subfigure[][]{\label{subfig:b6_selectivity}\includegraphics[width=0.05\textwidth]{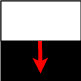}}\hspace{200pt} ~
	\subfigure[][]{\label{subfig:e6_selectivity}\includegraphics[width=0.05\textwidth]{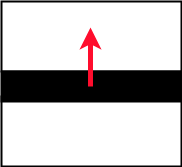}}\\
	\subfigure[]{\includegraphics[width=0.49\textwidth]{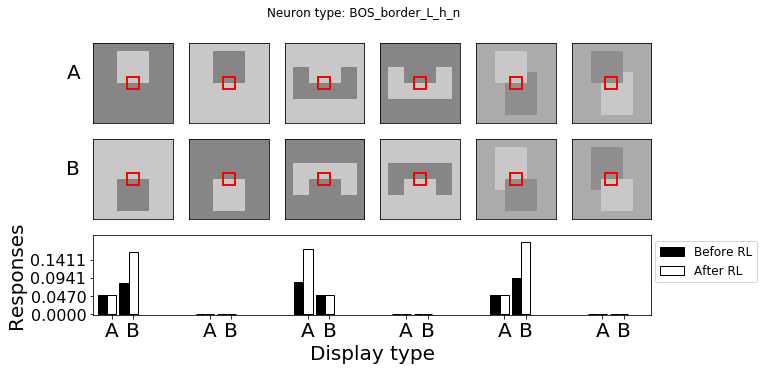}\label{subfig:border_L_h_n_resp}} ~
	\subfigure[]{\includegraphics[width=0.49\textwidth]{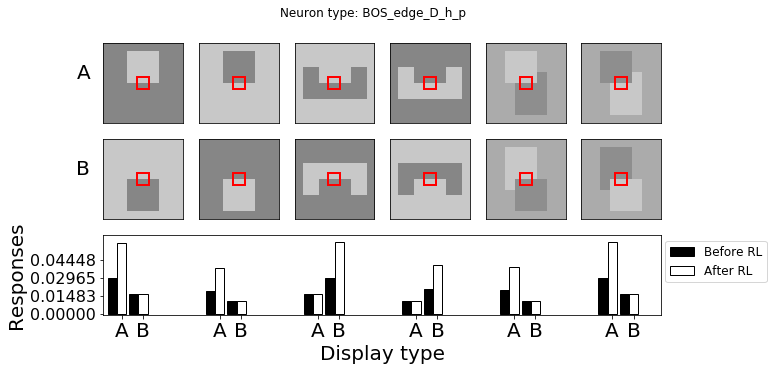}\label{subfig:edge_D_h_p_resp}}\\
	\subfigure[]{\includegraphics[width=0.49\textwidth]{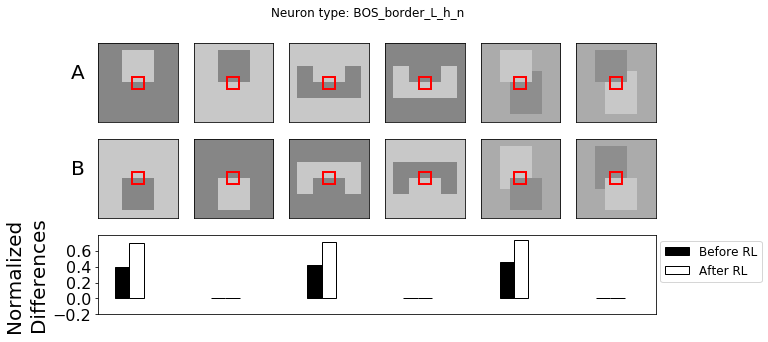}\label{subfig:b6_norm_diff_border}} ~
	\subfigure[]{\includegraphics[width=0.49\textwidth]{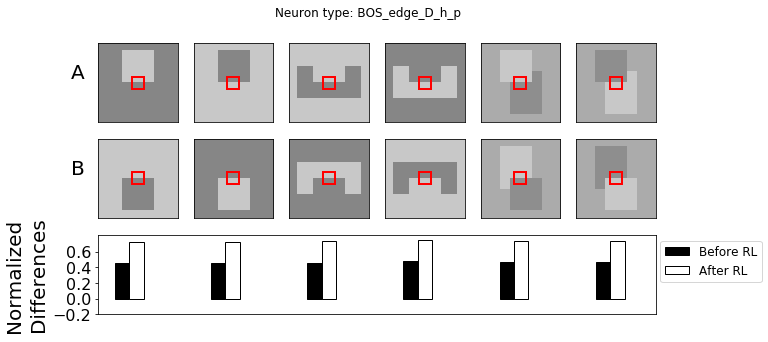}\label{subfig:e6_norm_diff_edge}}\\
	\subfigure[]{\includegraphics[width=0.45\textwidth]{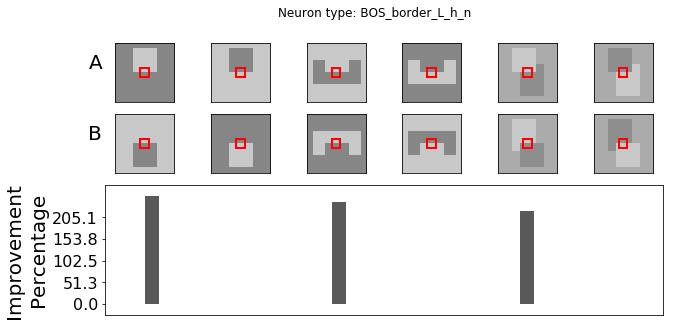}\label{subfig:b6_imp_perc_border}} \hspace{20pt}~
	\subfigure[]{\includegraphics[width=0.45\textwidth]{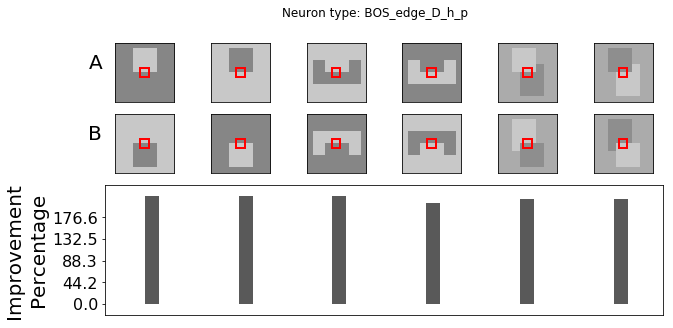}\label{subfig:e6_imp_perc_edge}}\\
	\caption{Example of model border ownership responses to stimuli used in the neurophysiological experiments by Zhou \etal \cite{BOwn_Zhou}. Note that the stimuli are rotated to match the orientation selectivity of the neuron. \subref{subfig:b6_selectivity} and \subref{subfig:e6_selectivity} represent the selectivity of neurons for which responses are shown in columns of this figure. The red arrows show the border ownership direction selectivity. In \subref{subfig:border_L_h_n_resp} and \subref{subfig:edge_D_h_p_resp}, dark bars indicate the initial border ownership responses obtained by MT modulations, and the light bars show the responses after relaxation labeling. In \subref{subfig:b6_norm_diff_border} and \subref{subfig:e6_norm_diff_edge}, the normalized differences, \ie, the relative amount of difference in responses to preferred and non-preferred stimuli are depicted. In \subref{subfig:b6_imp_perc_border} and \subref{subfig:e6_imp_perc_edge}, the percentage of improvement after relaxation labeling is shown.}
	\label{fig:BOS_model_responses6}
\end{figure}
\begin{figure}[]
	\centering
	\subfigure[][]{\label{subfig:b7_selectivity}\includegraphics[width=0.05\textwidth]{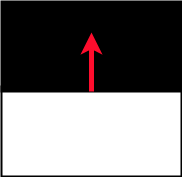}}\hspace{200pt} ~
	\subfigure[][]{\label{subfig:e7_selectivity}\includegraphics[width=0.05\textwidth]{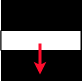}}\\
	\subfigure[]{\includegraphics[width=0.49\textwidth]{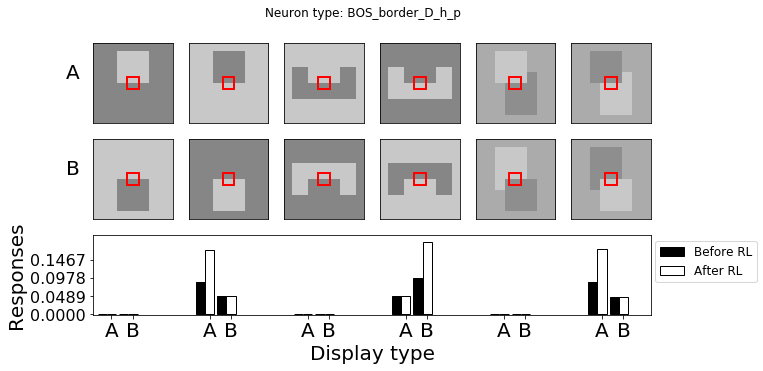}\label{subfig:border_D_h_p_resp}} ~
	\subfigure[]{\includegraphics[width=0.49\textwidth]{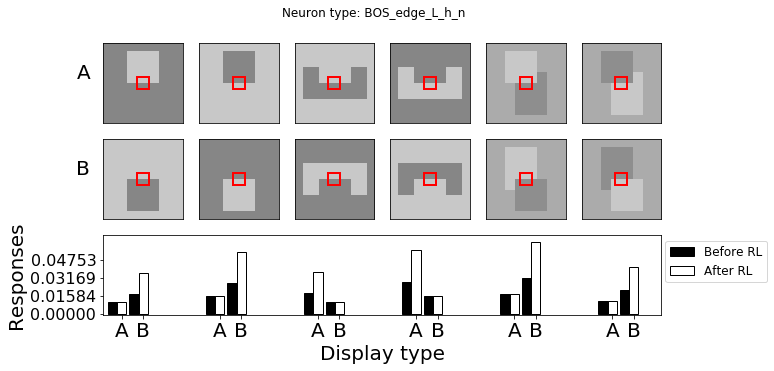}\label{subfig:edge_L_h_n_resp}}\\
	\subfigure[]{\includegraphics[width=0.49\textwidth]{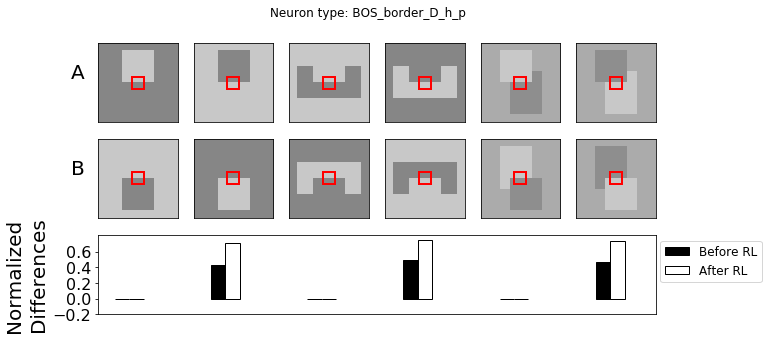}\label{subfig:b7_norm_diff_border}} ~
	\subfigure[]{\includegraphics[width=0.49\textwidth]{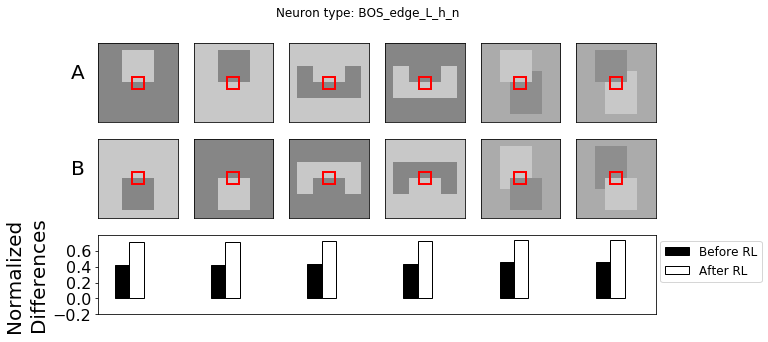}\label{subfig:e7_norm_diff_edge}}\\
	\subfigure[]{\includegraphics[width=0.45\textwidth]{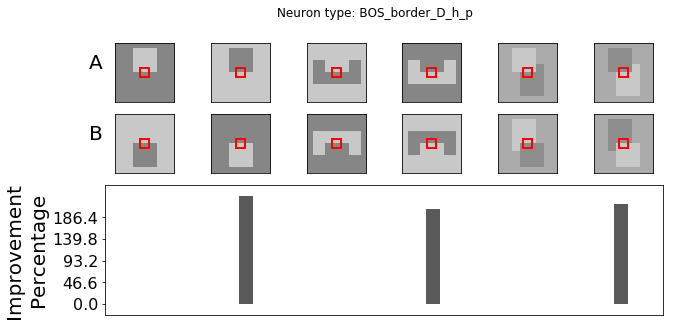}\label{subfig:b7_imp_perc_border}} \hspace{20pt}~
	\subfigure[]{\includegraphics[width=0.45\textwidth]{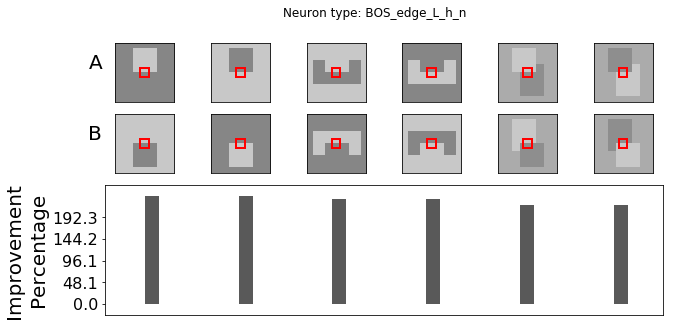}\label{subfig:e7_imp_perc_edge}}\\
	\caption{Example of model border ownership responses to stimuli used in the neurophysiological experiments by Zhou \etal \cite{BOwn_Zhou}. Note that the stimuli are rotated to match the orientation selectivity of the neuron. \subref{subfig:b7_selectivity} and \subref{subfig:e7_selectivity} represent the selectivity of neurons for which responses are shown in columns of this figure. The red arrows show the border ownership direction selectivity. In \subref{subfig:border_D_h_p_resp} and \subref{subfig:edge_L_h_n_resp}, dark bars indicate the initial border ownership responses obtained by MT modulations, and the light bars show the responses after relaxation labeling. In \subref{subfig:b7_norm_diff_border} and \subref{subfig:e7_norm_diff_edge}, the normalized differences, \ie, the relative amount of difference in responses to preferred and non-preferred stimuli are depicted. In \subref{subfig:b7_imp_perc_border} and \subref{subfig:e7_imp_perc_edge}, the percentage of improvement after relaxation labeling is shown.}
	\label{fig:BOS_model_responses7}
\end{figure}
\begin{figure}[]
	\centering
	\subfigure[][]{\label{subfig:b8_selectivity}\includegraphics[width=0.05\textwidth]{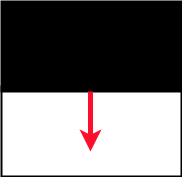}}\hspace{200pt} ~
	\subfigure[][]{\label{subfig:e8_selectivity}\includegraphics[width=0.05\textwidth]{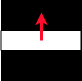}}\\
	\subfigure[]{\includegraphics[width=0.49\textwidth]{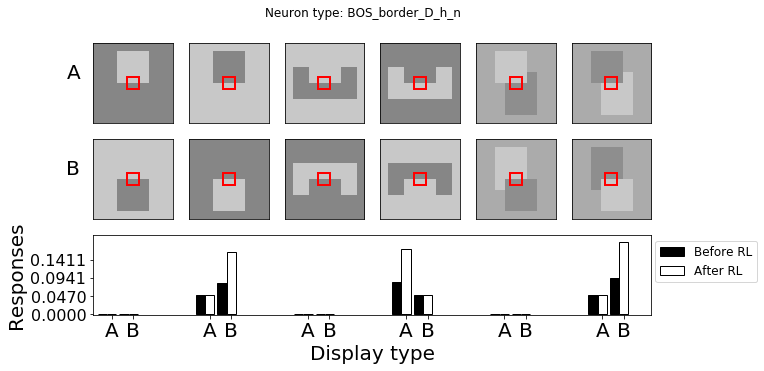}\label{subfig:border_D_h_n_resp}} ~
	\subfigure[]{\includegraphics[width=0.49\textwidth]{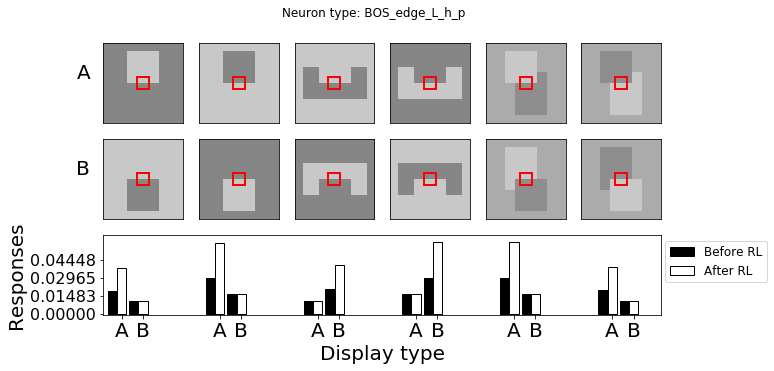}\label{subfig:edge_L_h_p_resp}}\\
	\subfigure[]{\includegraphics[width=0.49\textwidth]{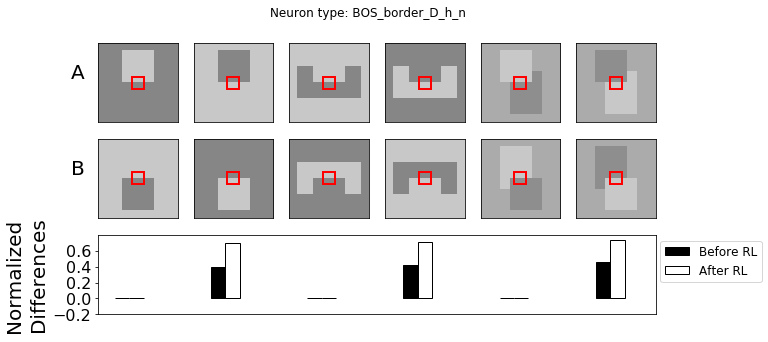}\label{subfig:b8_norm_diff_border}} ~
	\subfigure[]{\includegraphics[width=0.49\textwidth]{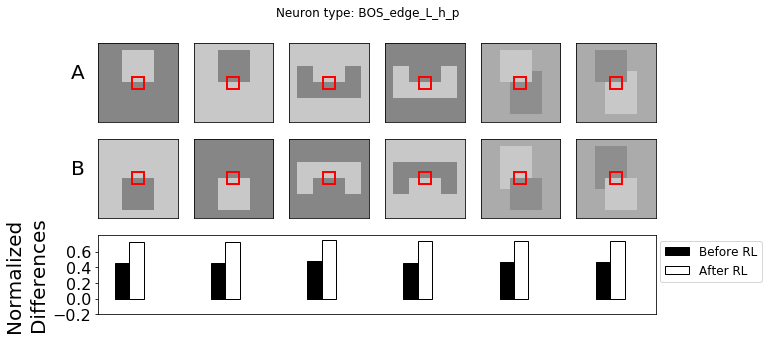}\label{subfig:b8_norm_diff_edge}}\\
	\subfigure[]{\includegraphics[width=0.45\textwidth]{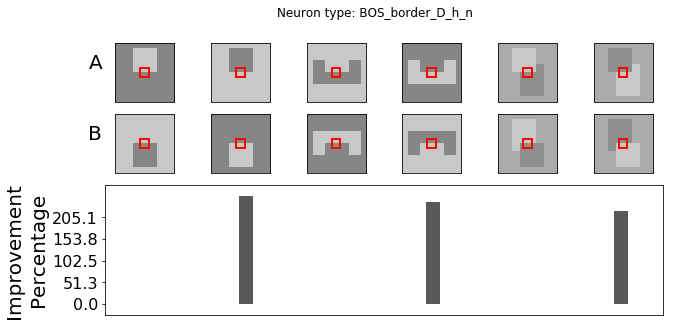}\label{subfig:b8_imp_perc_border}} \hspace{20pt}~
	\subfigure[]{\includegraphics[width=0.45\textwidth]{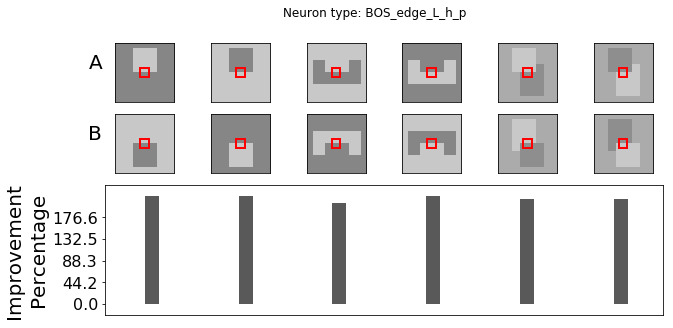}\label{subfig:b8_imp_perc_edge}}\\
	\caption{Example of model border ownership responses to stimuli used in the neurophysiological experiments by Zhou \etal \cite{BOwn_Zhou}. Note that the stimuli are rotated to match the orientation selectivity of the neuron. \subref{subfig:b8_selectivity} and \subref{subfig:e8_selectivity} represent the selectivity of neurons for which responses are shown in columns of this figure. The red arrows show the border ownership direction selectivity. In \subref{subfig:border_D_h_n_resp} and \subref{subfig:edge_L_h_p_resp}, dark bars indicate the initial border ownership responses obtained by MT modulations, and the light bars show the responses after relaxation labeling. In \subref{subfig:b8_norm_diff_border} and \subref{subfig:b8_norm_diff_edge}, the normalized differences, \ie, the relative amount of difference in responses to preferred and non-preferred stimuli are depicted. In \subref{subfig:b8_imp_perc_border} and \subref{subfig:b8_imp_perc_edge}, the percentage of improvement after relaxation labeling is shown.}
	\label{fig:BOS_model_responses8}
\end{figure}
\end{appendices}

\end{document}